\documentclass{article}

\PassOptionsToPackage{numbers, compress}{natbib}
     \usepackage[final,nonatbib]{neurips_2020}
\usepackage[sort,numbers]{natbib}

\usepackage[title]{appendix}
\usepackage[utf8]{inputenc} %
\usepackage[T1]{fontenc}    %
\usepackage[colorlinks]{hyperref}       %
\usepackage{url}            %
\usepackage{booktabs}       %
\usepackage{amsfonts}       %
\usepackage{nicefrac}       %
\usepackage{microtype}      %
\usepackage{graphicx}
\usepackage{calc}
\usepackage{subfigure}
\usepackage{booktabs} %
\usepackage{amsfonts}       %
\usepackage{nicefrac}       %
\usepackage{microtype}      %
\usepackage{xspace}
\usepackage{amsmath}
\usepackage{amssymb}
\usepackage{multirow}
\usepackage{bm}
\usepackage{cleveref}
\usepackage{xcolor}
\usepackage{pdflscape}
\usepackage{rotating}
\usepackage{caption}

\usepackage{pifont}%
\newcommand{\cmark}{\ding{51}}%

\newlength\myheight
\newlength\mydepth
\settototalheight\myheight{Xygp}
\newcommand*\inlinegraphics[1]{%
	\settototalheight\myheight{Xygp}%
	\settodepth\mydepth{Xygp}%
	\raisebox{-\mydepth}{\includegraphics[height=\myheight]{#1}}%
}

\newcommand{\bq}{\bm{q}}
\newcommand{\bk}{\bm{k}}

\newcommand{\bt}{\bm{t}}

\newcommand{\bv}{\bm{v}}
\newcommand{\bw}{\bm{w}}

\newcommand{\attnviz}[4]{%
\centering\vspace{10pt}
\addtolength{\tabcolsep}{-4pt}
\centerline{%
\begin{tabular}{c|ccc @{\hskip 5mm} c|ccc}
\multicolumn{1}{c}{} & \multicolumn{3}{c}{$7{\times} 7$ attention} & \multicolumn{1}{c}{} & \multicolumn{3}{c}{$14{\times} 14$ attention}\\
Query &  \multicolumn{3}{c}{Correspondence in support set} & Query &  \multicolumn{3}{c}{Correspondence in support set} \\
\includegraphics[height=#4]{supplementary/figs/attn-viz-split2/#1/q-0.jpg} & \includegraphics[height=#4]{supplementary/figs/attn-viz-split2/#1/s-0-0.jpg} & \includegraphics[height=#4]{supplementary/figs/attn-viz-split2/#1/s-0-1.jpg} & \includegraphics[height=#4]{supplementary/figs/attn-viz-split2/#1/s-0-2.jpg} &
\includegraphics[height=#4]{supplementary/figs/attn-viz-split2-14/#1/q-0.jpg} & \includegraphics[height=#4]{supplementary/figs/attn-viz-split2-14/#1/s-0-0.jpg} & \includegraphics[height=#4]{supplementary/figs/attn-viz-split2-14/#1/s-0-1.jpg} & \includegraphics[height=#4]{supplementary/figs/attn-viz-split2-14/#1/s-0-2.jpg} \\
\includegraphics[height=#4]{supplementary/figs/attn-viz-split2/#1/q-1.jpg} & \includegraphics[height=#4]{supplementary/figs/attn-viz-split2/#1/s-1-0.jpg} & \includegraphics[height=#4]{supplementary/figs/attn-viz-split2/#1/s-1-1.jpg} & \includegraphics[height=#4]{supplementary/figs/attn-viz-split2/#1/s-1-2.jpg} &
\includegraphics[height=#4]{supplementary/figs/attn-viz-split2-14/#1/q-1.jpg} & \includegraphics[height=#4]{supplementary/figs/attn-viz-split2-14/#1/s-1-0.jpg} & \includegraphics[height=#4]{supplementary/figs/attn-viz-split2-14/#1/s-1-1.jpg} & \includegraphics[height=#4]{supplementary/figs/attn-viz-split2-14/#1/s-1-2.jpg} \\
\includegraphics[height=#4]{supplementary/figs/attn-viz-split2/#1/q-2.jpg} & \includegraphics[height=#4]{supplementary/figs/attn-viz-split2/#1/s-2-0.jpg} & \includegraphics[height=#4]{supplementary/figs/attn-viz-split2/#1/s-2-1.jpg} & \includegraphics[height=#4]{supplementary/figs/attn-viz-split2/#1/s-2-2.jpg} &
\includegraphics[height=#4]{supplementary/figs/attn-viz-split2-14/#1/q-2.jpg} & \includegraphics[height=#4]{supplementary/figs/attn-viz-split2-14/#1/s-2-0.jpg} & \includegraphics[height=#4]{supplementary/figs/attn-viz-split2-14/#1/s-2-1.jpg} & \includegraphics[height=#4]{supplementary/figs/attn-viz-split2-14/#1/s-2-2.jpg} \\
\includegraphics[height=#4]{supplementary/figs/attn-viz-split2/#1/q-3.jpg} & \includegraphics[height=#4]{supplementary/figs/attn-viz-split2/#1/s-3-0.jpg} & \includegraphics[height=#4]{supplementary/figs/attn-viz-split2/#1/s-3-1.jpg} & \includegraphics[height=#4]{supplementary/figs/attn-viz-split2/#1/s-3-2.jpg} &
\includegraphics[height=#4]{supplementary/figs/attn-viz-split2-14/#1/q-3.jpg} & \includegraphics[height=#4]{supplementary/figs/attn-viz-split2-14/#1/s-3-0.jpg} & \includegraphics[height=#4]{supplementary/figs/attn-viz-split2-14/#1/s-3-1.jpg} & \includegraphics[height=#4]{supplementary/figs/attn-viz-split2-14/#1/s-3-2.jpg} \\
\includegraphics[height=#4]{supplementary/figs/attn-viz-split2/#1/q-4.jpg} & \includegraphics[height=#4]{supplementary/figs/attn-viz-split2/#1/s-4-0.jpg} & \includegraphics[height=#4]{supplementary/figs/attn-viz-split2/#1/s-4-1.jpg} & \includegraphics[height=#4]{supplementary/figs/attn-viz-split2/#1/s-4-2.jpg} &
\includegraphics[height=#4]{supplementary/figs/attn-viz-split2-14/#1/q-4.jpg} & \includegraphics[height=#4]{supplementary/figs/attn-viz-split2-14/#1/s-4-0.jpg} & \includegraphics[height=#4]{supplementary/figs/attn-viz-split2-14/#1/s-4-1.jpg} & \includegraphics[height=#4]{supplementary/figs/attn-viz-split2-14/#1/s-4-2.jpg} \\
\end{tabular}}
\vspace{-2mm}\captionof{figure}{\textbf{#2.} #3.}\label{#1}
\vspace{10pt}
\addtolength{\tabcolsep}{4pt}
\par}

\newif\ifarxiv
\newcommand{\refsupp}[1]{%
    \ifarxiv%
      {\Cref{#1}}%
    \else%
     {the extended version~\cite{doersch20}}%
    \fi%
}

\arxivtrue  %

\title{CrossTransformers: spatially-aware few-shot transfer}

\author{
Carl Doersch$^{*}$ \quad\quad Ankush Gupta$^{*}$ \quad\quad Andrew Zisserman$^{*\dagger}$ \vspace{1em} \\
  $^{*}$ DeepMind, London \quad\quad\quad\quad $^{\dagger}$ VGG, Department of Engineering Science, University of Oxford
}

\begin{document}

\maketitle

\begin{abstract}
Given new tasks with very little data---such as new classes in a classification problem or a domain shift in the input---performance of modern vision systems degrades remarkably quickly.
In this work, we illustrate how the neural network representations which underpin modern vision systems are subject to {\em supervision collapse}, whereby they lose any information that is not necessary for performing the training task, including information that may be necessary for transfer to new tasks or domains. 
We then propose two methods to mitigate this problem.
First, we employ self-supervised learning to encourage general-purpose features that transfer better.
Second, we propose a novel Transformer based neural network architecture called CrossTransformers, which can take a small number of labeled images and an unlabeled query, find coarse spatial correspondence between the query and the labeled images, and then infer class membership by computing distances between spatially-corresponding features.
The result is a classifier that is more robust to task and domain shift, which we demonstrate via state-of-the-art performance on Meta-Dataset, a recent dataset for evaluating transfer from ImageNet to many other vision datasets.
Code and pretrained checkpoints available at: \url{https://github.com/google-research/meta-dataset}.
\end{abstract}

\section{Introduction}
General-purpose vision systems must be adaptable.  
Home robots must be able to operate in new, unseen homes; photo-organizing software must recognize unseen objects (e.g., to find examples of ``my sixth-grade son's abstract art project''); industrial quality-assurance systems must spot defects in new products. 
Deep neural network representations can bring some visual knowledge from datasets like ImageNet~\cite{imagenet} to bear on different tasks beyond ImageNet~\cite{girshick14rcnn,oquab2014learning,Chatfield14a}, but empirically, this requires a non-trivial amount of labeled data in the new task.
With too little labeled data, or for a large change in distribution, 
such systems empirically perform poorly.

Research on meta-learning directly benchmarks adaptability.  
At training time, the algorithm receives a large amount of data and accompanying supervision (e.g., labels).  
At test time, however, the algorithm receives a series of {\em episodes}, each of which consists of a small number of datapoints from a {\em different distribution} than the training set (e.g., a different domain or different classes).  
Only a subset of this data has the accompanying supervision (called the {\em support set}); the algorithm must make predictions about the rest (the {\em query set}).  
Meta-Dataset~\cite{metadataset} is particularly relevant for vision, as the challenge is few-shot fine-grained image classification.  
The training data is a subset of ImageNet classes.
At test time, each episode either contains images from the other ImageNet classes, or from one of nine other visually distinct fine-grained recognition datasets.
The algorithm must rapidly adapt its representations to the new classes and domains.

\begin{figure}
  \centering
  \includegraphics[width=.91\textwidth]{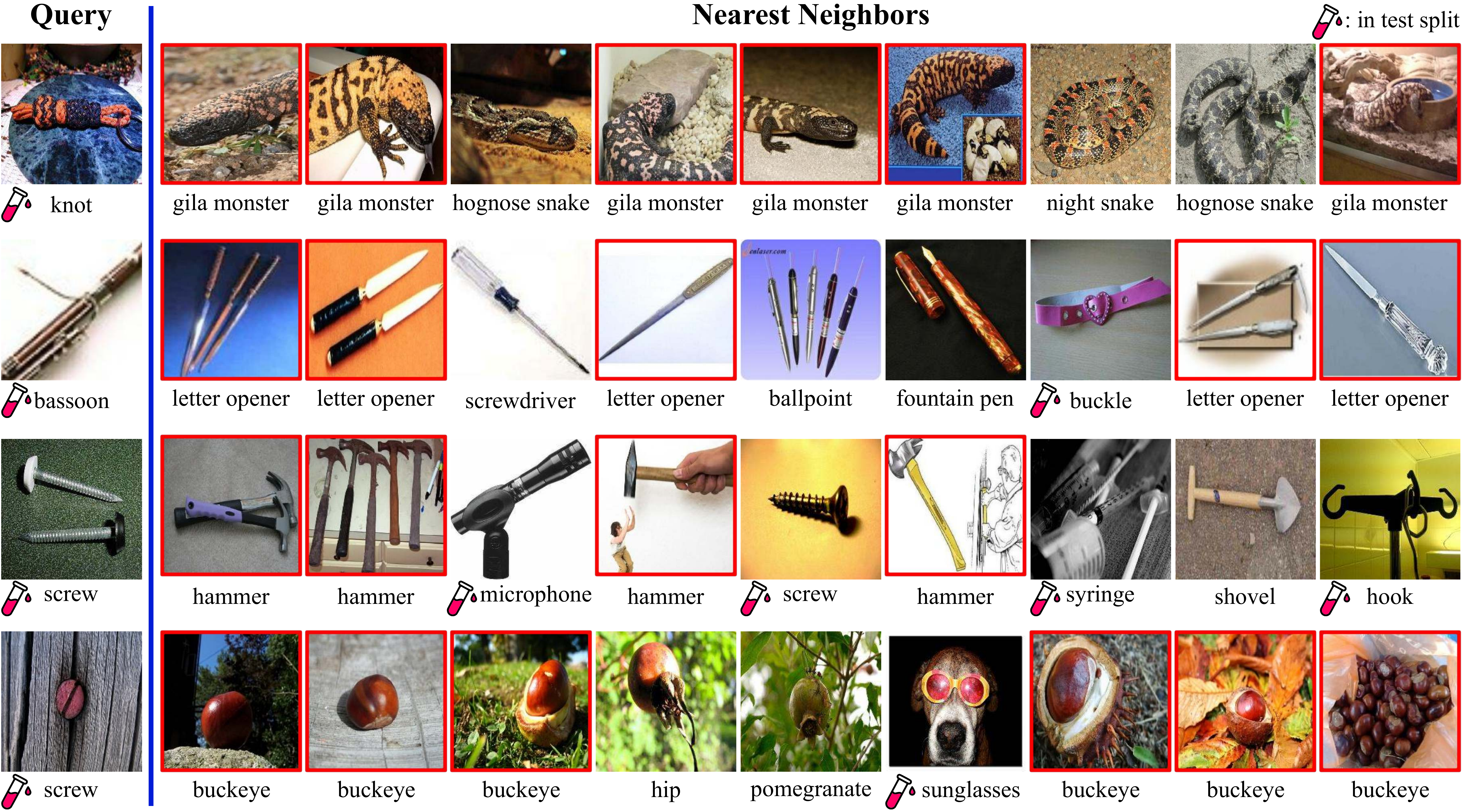}
  \caption{\textbf{Illustration of supervision collapse with nearest neighbors.}  
In each row, the leftmost image is a query taken from the Meta-Dataset ImageNet \emph{test} classes, and the rest are the top 9 nearest neighbors from \emph{both training and test} support set classes, using the embedding learned by a Prototypical Net 
(training details in \refsupp{a:nearest_neighbors}). Images belonging to the test split are indicated by a \protect\inlinegraphics{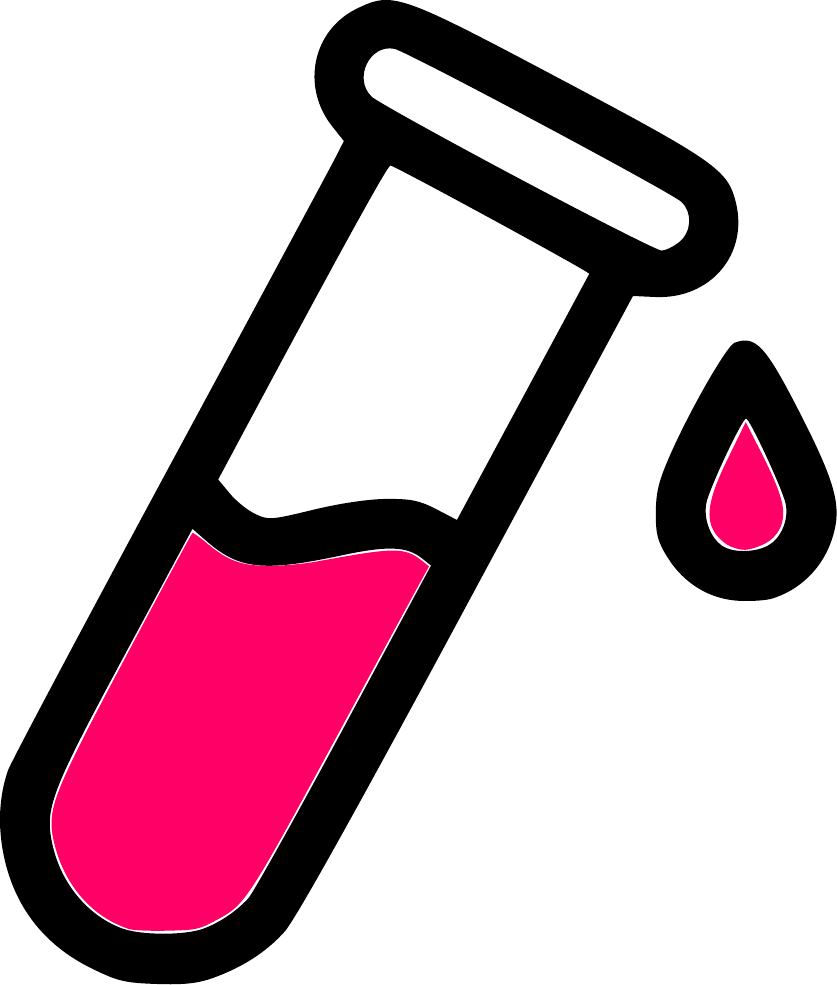} 
near the bottom left corner; rest are from the training split.
For a simple classifier to work well on these test classes, semantically similar images should have similar representations, and so we hope the nearest neighbors would come from the same---or semantically similar---classes.  
Instead, we observe that only ~5\% of matches for test-set queries are from the same class as the query.
Furthermore, many matches are all from the \emph{same incorrect training class} (highlighted in red).
We see a knot is matched with several gila monsters (and other reptiles); 
a bassoon with letter openers (and pens); 
a screw with hammers; another screw with buckeyes. 
The errors within that wrong class often have widely different appearance: for example, the bottom-most screw is matched with single buckeyes and also a pile of buckeyes.
One interpretation is that the network picks up on image patterns during 
training that allow images of each class to be  tightly grouped in the feature space, 
minimizing other ways that the image might be similar to other classes in preparation for a confident classification.
For out-of-domain samples, the network can then overemphasize a spurious image pattern  that suggests membership in one training-set class. This is the consequence of supervision collapse, where image patterns that might help make the correct associations are lost.
}\vspace{-10mm}\label{fig:category_collapse}
\end{figure}

Simple centroid-based algorithms like Prototypical Nets~\cite{protonet,chen2020new} are near state-of-the-art on Meta-Dataset, achieving around 50\% accuracy on the held-out ImageNet classes in Meta-Dataset's validation set (chance is roughly 1 in 20). An equivalent classifier trained on those validation classes can achieve roughly 84\% accuracy on the same challenge. What accounts for the enormous discrepancy between performance on within-distribution samples and out-of-distribution samples? 
We hypothesize that because the neural network backbone of Prototypical Nets is designed for classification, they do just this: represent \emph{only} an image's (training-set) class, and discard information that might help with out-of-distribution classes.
Doing so minimizes the losses for many meta-learning algorithms, including Prototypical Nets.
We call this problem {\em supervision collapse}, and illustrate it in \Cref{fig:category_collapse}. 

Our first contribution is to explore using self-supervision to overcome supervision collapse. 
We employ SimCLR~\cite{simclr}, which learns embeddings that discriminate between every image in the dataset while maintaining invariance to transformations (e.g., cropping and color shifts), thus capturing more than just classes.
However, rather than treat SimCLR as an auxiliary loss, we reformulate SimCLR as ``episodes'' that can be classified in the same manner as a training episode.

Our second contribution is a novel architecture called \emph{CrossTransformers}, which extends Transformers~\cite{transformers} to few-shot fine-grained classification. 
Our key insight is that objects and scenes are generally composed of smaller parts, with \emph{local} appearance that may be similar to what has been seen at training time. 
The classical example of this is the {\em centaur} that appeared in several early papers on visual representation~\cite{bronstein2009partial,jacobs2000class,veltkamp2001shape}, where the parts from the human and horse composed the centaur.

CrossTransformers operationalize this insight of (i) local part-based comparisons, 
and (ii) accounting for spatial alignment, resulting in a procedure for comparing
images which is more agnostic to the underlying classes.
In more detail, first a coarse alignment between geometric or functional parts in the query- and support-set images is established using attention as in Transformers. 
Then, given this alignment, distances between corresponding local features are 
computed to inform classification. We demonstrate this improves generalization to unseen classes and domains.

In summary, our contributions in this paper are:
(i) We improve the robustness of our local features with a self-supervised technique, modifying the state-of-the-art SimCLR~\cite{simclr} algorithm.
(ii) We propose the CrossTransformer, a network architecture that is spatially aware and performs few-shot classification using more local features, which improves transfer.
Finally, (iii) we evaluate and ablate how the choices in these algorithms impact Meta-Dataset~\cite{metadataset} performance, and demonstrate state-of-the-art results on nearly every dataset within it, often by large margins.

\section{Related Work}

\paragraph{Few-shot image classification.} Few-shot learning \cite{miller00oneshot,fei06one,lake15concept,hariharan17low} has recently been primarily addressed in the meta-learning framework~\cite{naik1992meta,schmidhuber1987evolutionary,thrun98learning}, where a model learns an update rule for the parameters of a base-learner model~\cite{bengio1992optimization,schmidhuber1992learning,bengio1990learning} through a sequence of training episodes~\cite{thrun98lifelong,vilalta02perspective}. The meta-learner either learns to produce new parameters directly from the new data \cite{schmidhuber1992learning,schmidhuber1993neural,perez2018film,ha2016hypernetworks,bertinetto2016learning,munkhdalai2017meta,rebuffi2017learning}, or learns to produce an update rule to iteratively optimize the base learner to fit the new data~\cite{hochreiter01l2l,ravi17,andrychowicz16learning,younger2001meta,bengio1992optimization,bertinetto2018meta}. \cite{maml,nichol18first,maclaurin2015gradient} do not use any explicit meta-learner model, but instead unroll the base-learner gradient updates and optimize for model initializations which generalize well on novel tasks.  
Matching-based methods~\cite{matchnet,protonet,relationnet,garcia2017few} instead learn representations for similarity functions~\cite{bromley1994signature,chopra2005learning,koch2015siamese,chen2020new,tian2020rethinking}, in the hope that the similarities will generalize to new data. CrossTransformers fall in this category, and share much of their architecture with Prototypical Nets~\cite{protonet}. 

\paragraph{Attention for few-shot learning.} CrossTransformers attend~\cite{bahdanau2014neural} individually over each class's support set to establish local correspondences, whereas Matching Networks~\cite{matchnet} attend over the whole support set to ``point to'' matching instances. 
\cite{mishra17snail} extend this idea to larger contexts using temporally dilated convolutions~\cite{oord2016wavenet}. In the limit, attention over long-term experiences accumulated in memories~\cite{sprechmann2018memory,santoro2016meta,kaiser2017learning,munkhdalai2017meta} can augment more traditional learning.

\paragraph{Correspondences for visual recognition.} CrossTransformers perform classification by matching more local parts. 
Discriminative
parts~\cite{bourdev2009poselets,doersch2012what,juneja2013blocks,han2015matchnet,Bart05,tokmakov2019learning}
and visual words~\cite{sivic2003video,zhang2007local} have a rich
history, and have found applications in deformable-parts
models~\cite{felzenszwalb2009object},
classification~\cite{zhang2007local,savarese2006discriminative},
and retrieval~\cite{cao2020unifying,tolias2020learning}.
Part-based correspondences for recognition~\cite{zheng2017learning}
have been particularly successful in fine-grained face retrieval and
recognition~\cite{xie18comparator,li2013probabilistic}.
CrossTransformers establish soft correspondences between pixels in the
query- and support-set images; such dense pairwise
interactions~\cite{wang2018non} have recently proved useful for
generative networks~\cite{zhang2019self}, semantic
matching~\cite{rocco2018neighbourhood} and tracking~\cite{vondrick18tracking}.
\cite{lifchitz2019dense} learns spatially dense classifiers for few-shot classification, but pools the spatial dimensions of the prototypes, and hence does not have a notion of correspondence.

\paragraph{Self-supervised learning for few-shot.} Our work on SimCLR episodes inherits from a line of self-supervised learning research, which typically deal with transfer from pretext tasks to semantic ones and must therefore represent more than their training data~\cite{dosovitskiy14disc,doersch15,noroozi2016unsupervised,zhang16color, larsson2016learning,zhang2017split,gidaris18rotation,simclr,moco,bachman19amdim,caron2018deep}.
Some recent works~\cite{gidaris19fs,su2019does} demonstrate that this can improve few-shot learning, although these use self-supervised auxiliary losses rather than integrating self-supervised instance discrimination~\cite{simclr,dosovitskiy14disc,wu2018unsupervised,tian2019contrastive,misra2020self} into episodic training.  
Also particularly relevant are methods that use self-supervision for correspondence \cite{jason2016back,vondrick18tracking,liu2019selflow}, which may in future work improve the correspondences that CrossTransformers use.

\section{Stopping Collapse: SimCLR Episodes and CrossTransformers}

We take a two-pronged approach to dealing with the supervision collapse problem.
Modern approaches to few-shot learning typically involve learning an embedding for each image, followed by a classifier that aggregates information across an episode's support set in order to classify the episode's queries. 
Our first step aims to use self-supervised learning to improve the embedding so it expresses information beyond the classes, in a way that is as algorithm-agnostic as possible.
Once we have these embeddings, we build a classifier called a CrossTransformer. 
CrossTransformers use Prototypical Nets~\cite{protonet} as a blueprint, chosen due to their simplicity and strong performance; the main modification is to aggregate information in a spatially-aware way. 
We begin by reviewing Prototypical Nets, and then describe the two approaches.

Prototypical Nets are \emph{episodic learners}, which means training is performed on the same kind of episodes that will be presented at test time: a query set $Q$ of images, and a support set $S$ which can be partitioned into classes $c \in \{1, 2, \hdots, C\}$: each $S^c = \{x_i^c\}_{i=1}^N$ is composed of $N$ example images $x_i^c$.
Prototypical Nets learn a distance function between the query and each subset $S^c$.
Both the query- and support-set images are first encoded into a $D$-dimensional representation $\Phi(x)$, using a shared ConvNet $\Phi: \mathbb{R}^{H\times W\times 3} \mapsto \mathbb{R}^{D}$, where $H, W$  are the height and width respectively. Then a ``prototype'' $\bt^c \in \mathbb{R}^{D}$ for the class $c$ is obtained by averaging the representations of the support set $S^c$, $\bt^c = \frac{1}{|S^c|} \sum_{x\in S^c} \Phi(x)$. Finally, a distribution of classes is obtained using softmax over the distances between the query image and class prototypes: $p(y = c | x_q) = \frac{\exp(-d(\Phi(x_q), \bt^c))}{\sum_{c'=1}^C \exp(-d(\Phi(x_q), \bt^{c'}))}$. In practice, the distance function $d$ is fixed to be the squared Euclidean distance $d(x_q, S^c) = ||\Phi(x_q) - \bt^c||^2_2$.
The learning objective is to train the embedding network $\Phi$ to maximize the probability of the correct class for each query. 

\subsection{Self-supervised training with SimCLR}\label{s:simclr}
Our first challenge is to improve the neural network embedding $\Phi$: after all, if these features have collapsed to represent little information beyond the classes, 
then a subsequent classifier cannot can recover this information.
But how can we train features to represent things beyond labels when our only supervision is the labels?
Our solution is self-supervised learning, which invents ``pretext tasks'' that train representations \emph{without} labels~\cite{doersch15,dosovitskiy14disc}, and better yet, has a reputation for representations that transfer beyond this pretext task.
Specifically we use SimCLR~\cite{simclr}, which uses ``instance discrimination'' as a pretext task.
It works by applying random image transformations (e.g., cropping or color shifts) twice to the same image, thus generating two ``views'' of that image.
Then it trains the network so that representations of the two views of the same image are more similar to each other than they are to those of different images.
Empirically, networks trained in this way become sensitive to semantic information, but also learn to discriminate between different images \emph{within a single class}, which is useful for combating supervision collapse.

While we could treat SimCLR as an auxiliary loss on the embedding, we instead reformulate SimCLR as episodic learning, so that the technique can be applied to all episodic learners with minimal hyper-parameters.
To do this, we randomly convert 50\% of the training episodes into what we call \emph{SimCLR episodes}, by treating every image as its own class.
For clarity, we will call the original episodes that have not been converted SimCLR episodes \emph{MD-categorization episodes}, to emphasize that they use the original categories from Meta-Dataset.
Specifically, let $\rho(\cdot)$ be SimCLR's (random) image transformation function, and let $S=\{x_i\}^{|S|}_{i=1}$ be a training support set.
We generate a SimCLR episode by sampling a new support set, transforming each image in the original support set $S^{\prime}=\{\rho(x_i)\}^{|S|}_{i=1}$, and then generating query images by sampling other transformations from the same support set: $Q^{\prime}=\{\rho(\operatorname{random\_sample}(S))\}^{|Q|}_{i=1}$, where $\operatorname{random\_sample}$ just takes a random image from the set.\footnote{We enforce that the sampled queries have the same class distribution as $Q$, and have no repeats.}
The original query set $Q$ is discarded.
The label for an image in the SimCLR episode is its index in the original support set, resulting in an $|S|$-way classification for each query. 
Note that for a SimCLR episode, the `prototypes' in Prototypical Nets average over just a single image, and therefore the Prototypical Net loss can be written as $\frac{\exp(-d(\Phi(\rho(x_q)), \Phi(\rho(x_q))))}{\sum_{i=1}^n \exp(-d(\Phi(\rho(x_q)), \Phi(\rho(x_i)))}$.
If we define $d$ as the cosine distance rather than Euclidean, this loss is identical to the one used in SimCLR.

\subsection{CrossTransformers}

\begin{figure}
    \centering
    \includegraphics[width=.99\textwidth]{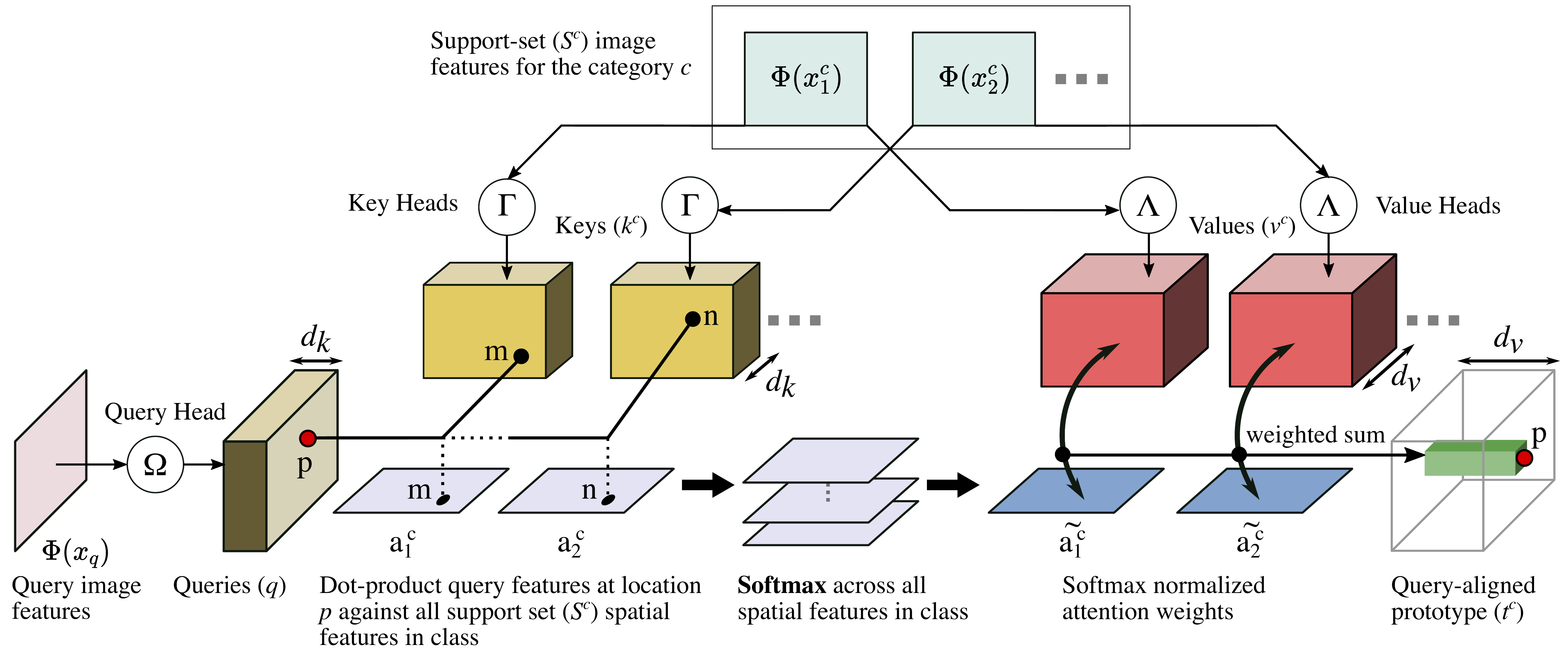}
    \caption{\textbf{CrossTransformers.} Construction of \emph{query-aligned}
    class prototype vector $t^c_p$ for the class $c$ and the query image
    $x_q$, focusing on the spatial location $p$ in $x_q$.
    The query vector $q_p$ is 
    compared against keys $k^c$ from all spatial locations in the support set 
    $S^c$ to obtain attention scores $a^c$, which are softmax normalized 
    before being used to aggregate the values $v^c$ for the aligned prototype 
    vector $t^c_p$.}
    \label{fig:crosstransformer}
    \vspace{-1em}
\end{figure}

Given a query image $x_q$ and a support set $S^c = \{x^c_i\}_{i=1}^N$ for 
the class $c$, CrossTransformers aim to build a representation which enables
local part-based comparisons between them.

CrossTransformers start by making the image representation a spatial tensor, and 
then assemble \emph{query-aligned} class prototypes
by putting the support-set images $S^c$ in correspondence with the query image.
The distance between the query image and the query-aligned  prototype for each 
 is then
computed and used in a similar way to Prototypical Nets.
In practice, we establish soft correspondences using 
attention~\cite{bahdanau2014neural} based Transformers~\cite{transformers}.
In contrast, Prototypical Nets use flat vector representations which lose the 
location of image features, and have a fixed class prototype which is
independent of the query image.

\begin{figure}
  \centering
  \addtolength{\tabcolsep}{-4pt}
  \centerline{%
  \resizebox{\linewidth}{!}{\begin{tabular}{c|ccc @{\hskip 5mm} c|ccc}
  Query &  \multicolumn{3}{c}{Correspondence in support set} & Query &  \multicolumn{3}{c}{Correspondence in support set} \\
  \includegraphics[height=20mm]{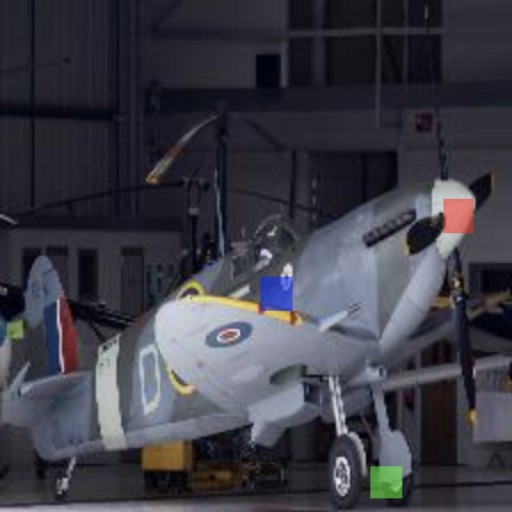} & \includegraphics[height=20mm]{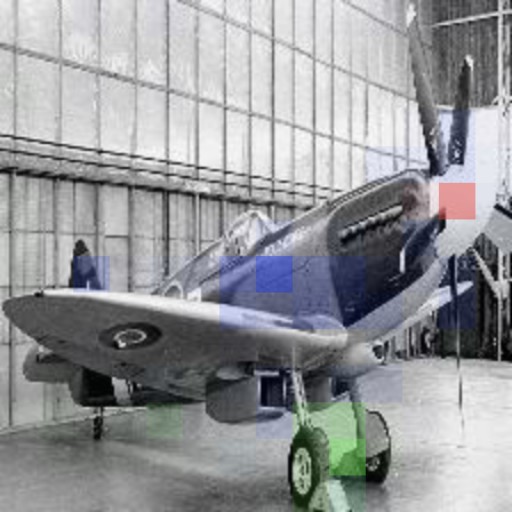} & \includegraphics[height=20mm]{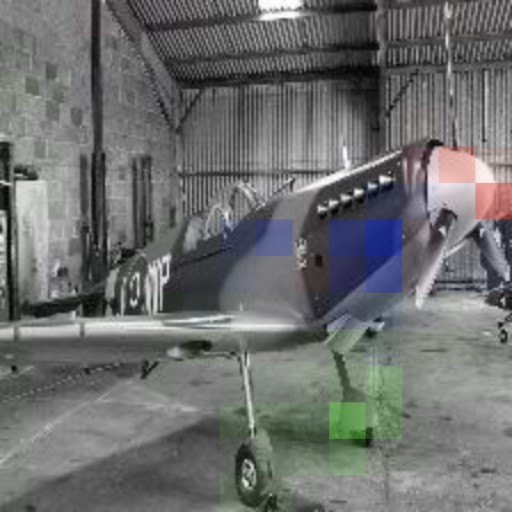} & \includegraphics[height=20mm]{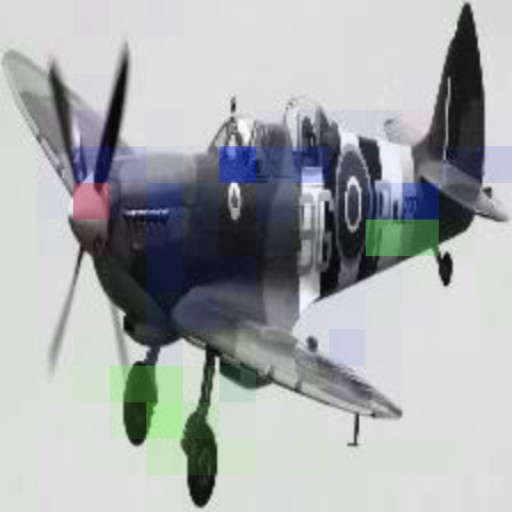} &
  \includegraphics[height=20mm]{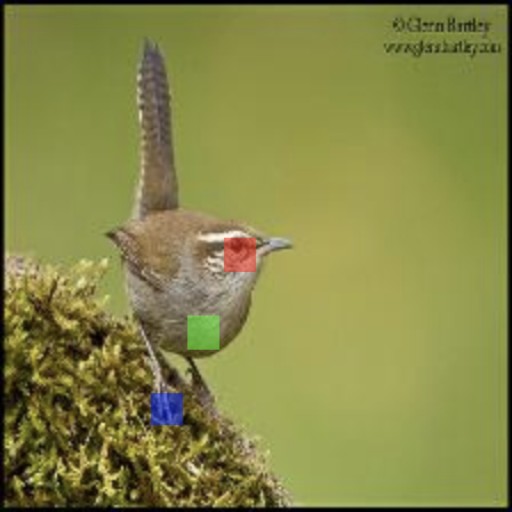} & \includegraphics[height=20mm]{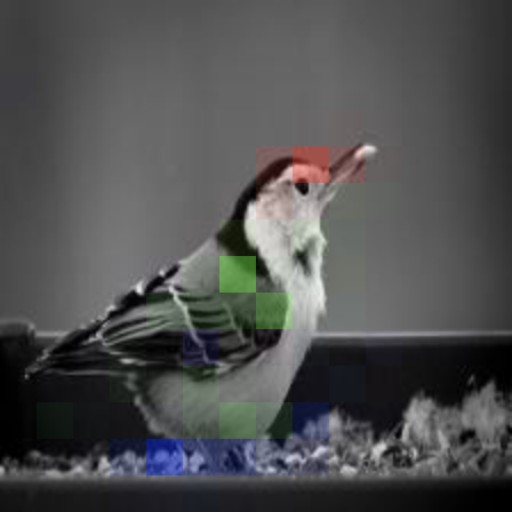} & \includegraphics[height=20mm]{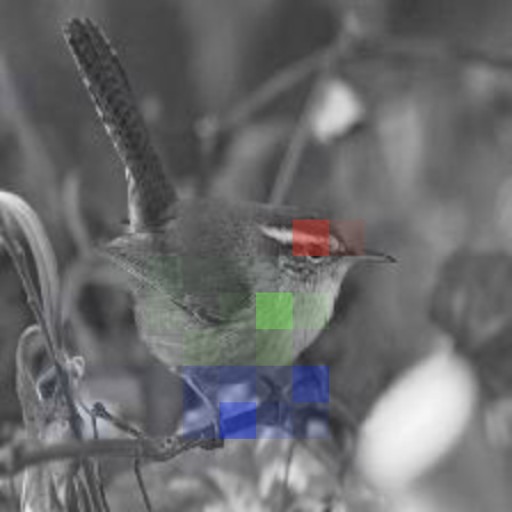} & \includegraphics[height=20mm]{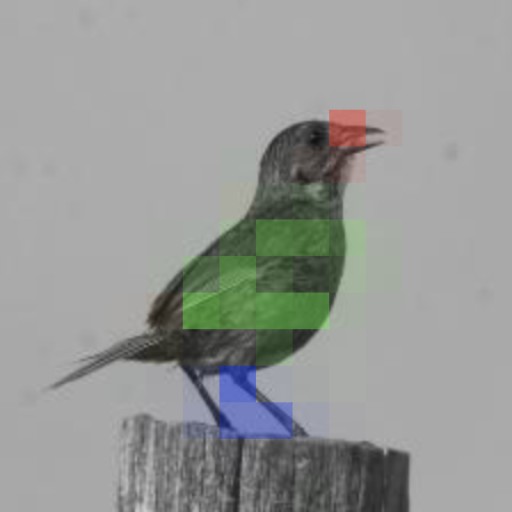}\\
  \includegraphics[height=20mm]{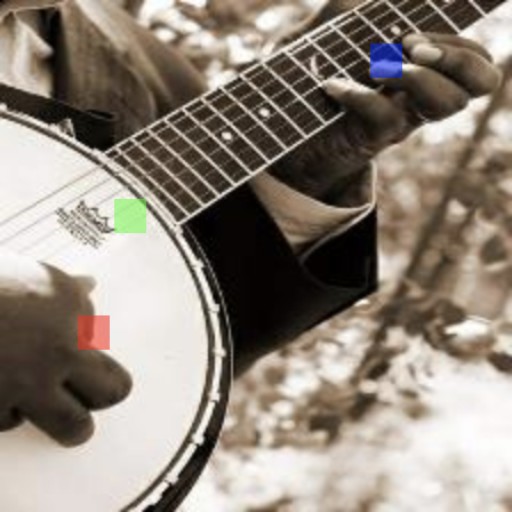} & \includegraphics[height=20mm]{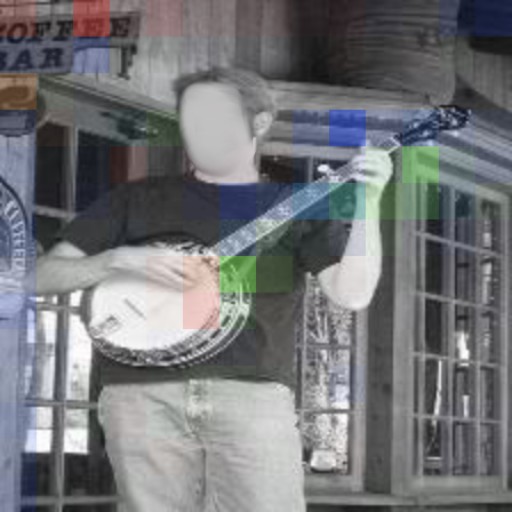} & \includegraphics[height=20mm]{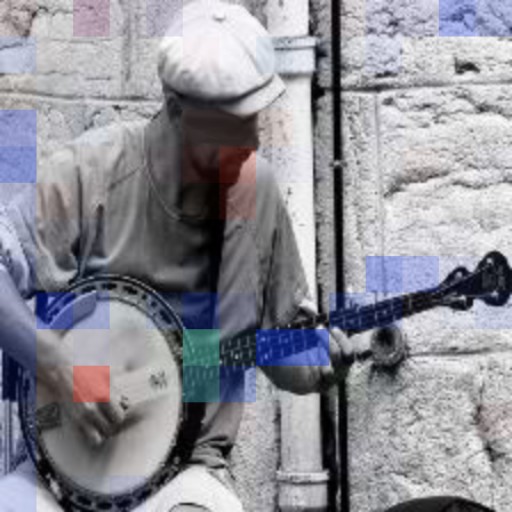} & \includegraphics[height=20mm]{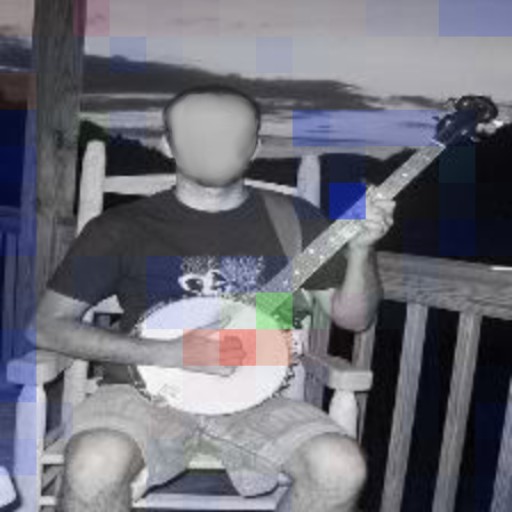} &
  \includegraphics[height=20mm]{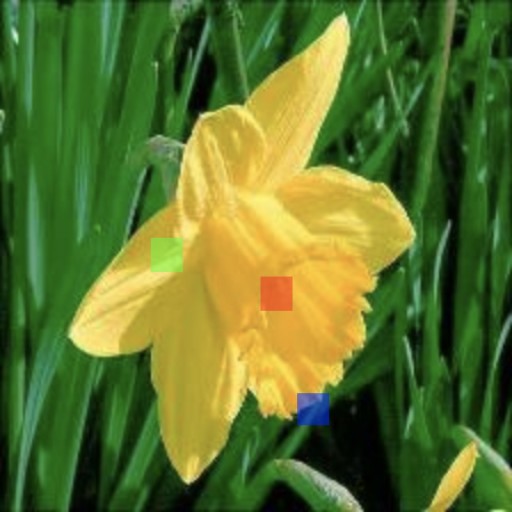} & \includegraphics[height=20mm]{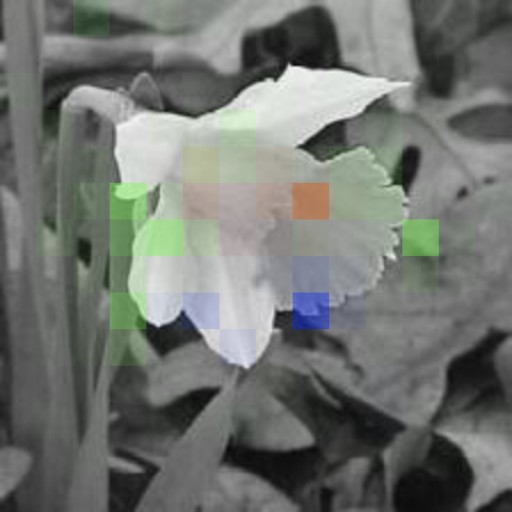} & \includegraphics[height=20mm]{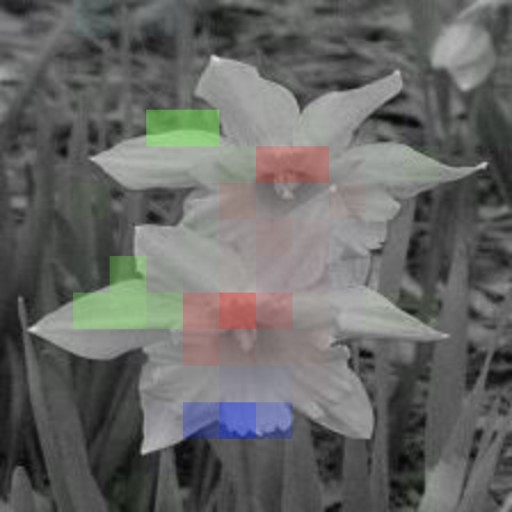} & \includegraphics[height=20mm]{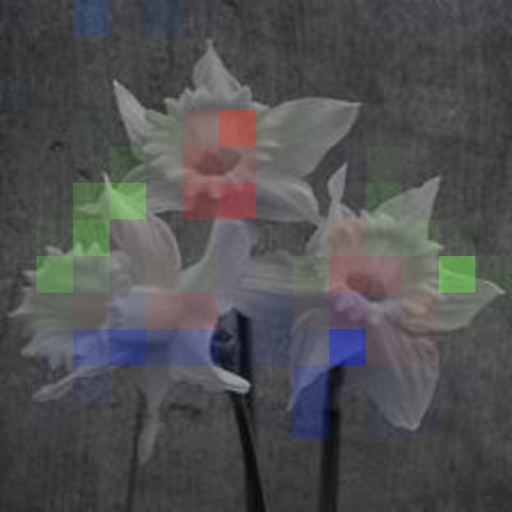}
  \end{tabular}}}
  \caption{
  \textbf{Visualization of the attention $\tilde{a}$.}
  We show four query images, along with three support-set images for each.
  Within each query image, we choose three spatial locations (red, green, and blue squares), and plot the CrossTransformer attention weights for each one in the corresponding color (brighter colors mean higher weight).
  The four examples are from Aircraft, CU-Birds, VGG Flowers, and ImNet test sets respectively (clockwise, starting from top-left).
  No matter which dataset, the attention masks are semantically relevant, even when the correspondence is not one-to-one.
  More visualizations are given in \refsupp{a:corrviz}.}

  \label{fig:attention_viz}
\end{figure}

Concretely, CrossTransformers remove the final spatial pooling in Prototypical 
Nets' embedding network $\Phi(\cdot)$, such that the spatial dimensions $H', W'$
are preserved: $\Phi(x) \in \mathbb{R}^{H'\times W'\times D}$. 
Following Transformers, key-value pairs are then generated for each image in the
support set using two independent linear maps: the key-head 
$\Gamma: \mathbb{R}^{D} \mapsto \mathbb{R}^{d_k}$, and the value-head 
$\Lambda: \mathbb{R}^{D} \mapsto \mathbb{R}^{d_v}$ respectively. 
Similarly, the query image features $\Phi(x_q)$ are embedded using the 
query-head $\Omega: \mathbb{R}^{D} \mapsto \mathbb{R}^{d_k}$.
Dot-product attention scores are then obtained between keys and queries,
followed by softmax normalization across all the images and locations in $S^c$.
This attention serves as our coarse correspondence (see example attention 
visualizations in \Cref{fig:attention_viz} and \refsupp{a:corrviz}),
and is used to aggregate the support-set features into alignment with the query. 
This process is visualized in \Cref{fig:crosstransformer}.

Mathematically, let $\bk^c_{jm} = \Gamma\cdot\Phi(x^c_j)_{m}$ be the key for the 
$j^{\text{th}}$ image in the support set for class $c$ at spatial position $m$ 
(index over the two dimensions $H',W'$), and similarly let
$\bq_p = \Omega\cdot\Phi(x_q)_{p}$ be the query vector at spatial position 
$p$ in the query image $x_q$.
The attention 
$\tilde{a}^{c}_{jmp} \in \mathbb{R}$ between the two is then obtained as:

\vspace{-.3em}
\begin{equation}
    \tilde{a}^{c}_{jmp}=\frac{\exp(a^c_{jmp}/\tau)}{\sum_{i,n}\exp(a^c_{inp}/\tau)},
    \qquad \text{where} \quad
    a^c_{jmp}=\bk^c_{jm} \cdot \bq_p,
    \quad \text{and} \quad \tau = \sqrt{d_k}.
\end{equation}
\vspace{-.3em}

Next, the aligned prototype vector $\bt^c_{p}$ corresponding to spatial location 
$p$ in the query is obtained by aggregating the support-set values $\bv^c_{jm} = \Lambda\cdot\Phi(x^c_j)_{m}$ using the attention weights above:

\vspace{-.3em}
\begin{equation}
    \bt^c_{p} = \sum_{jm} \tilde{a}^{c}_{jmp} \bv^{c}_{jm}
\end{equation}

\vspace{-.3em}
Finally, squared Euclidean distances between aligned local features from the 
above prototype and corresponding query image values $\bw_p = \Lambda\cdot\Phi(x^q)_{p}$ are aggregated as below.
This scalar distance acts as a negative logit for a distribution over classes as in Prototypical Nets.

\vspace{-.3em}
\begin{equation}
    d(x_q, S^c) = \frac{1}{H^{\prime}W^{\prime}}\sum_{p} ||\bt^c_{p}-\bw_{p}||_2^2
\end{equation}
\vspace{-.3em}

Note we apply the same value-head $\Lambda$ to both the query and support-set images. 
This ensures that the CrossTransformer behaves somewhat like a distance.
That is, imagine a trivial case where, for one class, all images in $S^c$ are identical to $x_q$. 
We would want $d(x_q, S^c)$ to approach 0 even if the network is untrained, or if these images are highly dissimilar from those used for training.
Sharing $\Lambda$ between the support and query sets helps accomplish this: in fact, if $\tilde{a}^{c}_{jmp}$ is 1 where $p=m$ and 0 elsewhere for all $j$, then $d(x_q, S^c)$ will be identically 0 under this architecture, no matter the network weights. 
To encourage this behavior for the attention $\tilde{a}$, we also set $\Gamma = \Omega$, i.e., the key 
and query heads are the same. This way, in our trivial case, the attention is likely to be 
maximal for spatial locations that correspond, because $\bk^c_{jm}$ and $\bq_p$ will be the same for $p=m$. 

For one experiment, we also augment the CrossTransformer with a global feature, which can help for some datasets like DTD (Describable Textures Dataset) with less spatial structure.

\section{Experiments}

We evaluate on Meta-Dataset~\cite{metadataset}, specifically the setting where the training is performed on the ImageNet train split only, which is 712 classes (plus 158 classes for validation, which are not used for training but only to perform early stopping).
We then test on the remaining 130 held-out classes from ImageNet, as well as 9 other image datasets.
Note that this is in contrast to another popular (and easier) setting, where the training also uses a subset of categories from more of these datasets: usually all datasets except Traffic Signs and COCO.
For clarity, we'll use ``Meta-Dataset Train-on-ILSVRC'' to denote training on ImageNet only, and ``Meta-Dataset Train-on-all'' to denote when training occurs on more datasets.
Test time consists of a series of episodes, each of which contains: (1) a support set between 50 and 500 labeled images which come from between 5 and 50 classes; and (2) an unlabeled query set with 10 images per class.
Meta-Dataset aims for fine-grained recognition, so the classes in each episode are mutually similar: one episode may contain only musical instruments, another may contain only birds, etc. 

Meta-Dataset is useful for studying transfer because different test datasets encapsulate different  kinds of transfer
challenges.
For test datasets like CU-Birds~\cite{wah2011cubirds}, there are numerous similar classes in ImageNet train  (20 bird classes in ImageNet train, versus 100 in the CU-birds test dataset).
In contrast, for test datasets like Aircraft~\cite{maji2013airplanes}, there is just a single corresponding class in ImageNet train; therefore, algorithms which don't represent the intra-class variability for this class will be penalized.
The ImageNet test set has images in a similar domain to  the ImageNet train set but with different classes, while test datasets like COCO 
contain many similar classes to  ImageNet, but with domain shift (in COCO, instances are generally not the subject of their photographs, and may be low-resolution or occluded).
Finally, test datasets like OmniGlot combine these challenges, i.e., different classes in a substantially different domain.

\subsection{Implementation details}

To ensure comparability, we followed the public implementation of Prototypical Nets for Meta-Dataset~\cite{metadataset} wherever possible.  
This includes using the same hyperparameters, unless otherwise noted.
For the hyperparameters that were chosen with a sweep on the validation set (learning rate schedule and weight decay), we simply used the best values discovered for Prototypical Nets for all the experiments in this paper.
See \refsupp{a:impl:ctx} for details of the CrossTransformer architecture.
We use no pretraining for CrossTransformers, although to be consistent with prior work~\cite{metadataset} we use it for the experiments involving Prototypical Nets.

We incorporate two improvements from Meta-Baseline~\cite{chen2020new}, which at test time is similar to Prototypical Nets (though it isn't trained as an episodic learner).
The first is to keep exponential moving averages for Batch Norm statistics during training, and use those for Batch Norm at test time.
Second, we note that Meta-Baseline does not train on fine-grained episodes sampled from the ImageNet hierarchy, as Prototypical Nets does, but rather on batches with uniformly-sampled classes.
Empirically, Prototypical Nets trained only on fine-grained episodes struggle to do coarse-grained recognition, as required for datasets like COCO.
Therefore, we only use the ImageNet hierarchy to make 50\% of episodes fine-grained; the rest have categories sampled uniformly.

\paragraph{Choice of network.}
Prior implementations of networks like Prototypical Nets use relatively small networks (e.g., ResNet-18) with small input images (e.g. 126$\times$126 pixels), and report that measures to increase capacity (e.g., Wide ResNets~\cite{Zagoruyko2016WRN}) provide minimal benefits.
This is surprising given that standard networks show improvements for increasing capacity (e.g.,  ResNet-34 outperforms ResNet-18 by 3\% on ImageNet~\cite{he2016deep}). Making our networks spatially-aware requires higher-resolution, and also higher-capacity networks are especially important in self-supervised learning~\cite{doersch17multi,kolesnikov2019revisiting}.
Therefore, our experiments increase resolution to the standard 224$\times$224 and use ResNet-34, and we also use normalized stochastic gradient descent~\cite{cortes2006finite,nesterov1984minimization}, which we found improved stability when fine-tuning more complex networks.
Table~\ref{tab:simclr-episodes} compares the 
Prototypical Nets performance of this network to that of using a ResNet-18.
Increased capacity leads to only slight performance improvements, which are more pronounced for datasets that are similar to ImageNet; it harms, e.g., OmniGlot.  
Further details in \refsupp{a:impl:train}.

For experiments with CrossTransformers, we also increased the resolution of the convolutional feature map by setting the stride of final block of the ResNet to 1, and using dilated convolutions to preserve the feature computation~\cite{giusti2013fast,holschneider1990real}.
This turns the usual 7$\times$7 feature map for a 224$\times$224 image into a 14$\times$14 feature map.
We ablate this choice in \refsupp{a:impl:ctx}.

\paragraph{Augmenting CTX with a global feature.}
Recent works have also shown benefits for applying logistic regression (LR) at test time~\cite{tian2020rethinking}.  
In practice, it is too expensive to apply LR to our query-aligned prototypes (as this would involve a separate classifier for every query).
Therefore, we instead apply logistic regression to a globally-pooled feature and average the logits with those produced by the CrossTransformer.
See \refsupp{a:impl:global} for details.

\paragraph{Augmentation.} While most experiments use no augmentation (apart from SimCLR episodes) to be consistent with prior work~\cite{metadataset}, more recent work~\cite{saikia2020optimized,chen2020new,tian2020rethinking} showed that stronger data augmentation is effective.
Therefore, for two experiments, we employ augmentation using the settings discovered in BOHB~\cite{saikia2020optimized} (via Auto-Augment~\cite{autoaugment} on the validation set), with an extra stage that randomly downsamples and then upsamples images, which we find helpful as our network operates at higher resolution than many of the test datasets.
This BOHB augmentation is only applied to the ``MD-categorization'' episodes, and not to the SimCLR episodes. 
Note this BOHB augmentation is different from SimCLR-style augmentation, which is used in SimCLR Episodes as well as in the ablation (SC-Aug) in \Cref{tab:simclr-episodes}.
See \refsupp{a:impl:aug} for details.

\subsection{Results for self-supervised learning with SimCLR on Prototypical Nets}\label{sec:self-supervised}

We first analyze the impact of SimCLR Episodes and other architectural choices in \Cref{tab:simclr-episodes}.
For baseline Prototypical Nets, SimCLR Episodes generally improve performance, but this depends on architectural choices.
Improvements are largest for datasets that are more distant from ImageNet, e.g., OmniGlot and Quickdraw, and datasets which require distinguishing between sub-categories ImageNet categories, e.g., Aircraft and Traffic Signs.
In ImageNet, all commercial airplanes fall in a single ImageNet class; therefore, the success of SimCLR Episodes here suggests they recover features which are lost due to supervision collapse.
Strangely, however, SimCLR Episodes interact with Batch Norm: we find more robust improvements when computing Batch Norm statistics from the test-time support set, but not when using exponential moving averages (EMA) as suggested by \cite{chen2020new}.  
One possible interpretation is that the network has learned to use Batch Norm to communicate information across the batch: e.g., to distinguish between SimCLR Episodes and MD-categorization episodes.
Using EMA at test time may prevent this, which may confuse the network.  
Interestingly, we will show later that SimCLR Episodes don't harm CrossTransformers as they harm Prototypical Nets when using EMA at test time, suggesting the two architectures solve the problem differently.

Recall that converting an MD-categorization episode into a SimCLR episode makes two changes to the episode: it 1) applies data augmentation, and 2) converts the classification problem to ``instance discrimination,'' by selecting images from the support set as a new query set, and requiring the network to predict the selected indices.
To ensure that we are not simply seeing the effect of data augmentation, we also implemented a baseline (SC-Aug) that does 1 but not 2 to the input MD-categorization episodes, and does this augmentation for all episodes (rather than 50\%, which is the fraction of MD-categorization episodes that are converted to SimCLR episodes for SC-Eps experiments).
Indeed, we see no improvements for this change, and in fact non-trivial performance loss from this, mirroring the result for supervised learning in the original paper~\cite{simclr}.
This reinforces that SimCLR was designed for self-supervised learning, and so the transformations are more severe than is usually optimal for supervised learning.

Finally, we see small improvements from using larger networks and higher resolution for the baseline model.  
While our baseline is overall better than the baseline Prototypical Nets implementation~\cite{metadataset}, it is still below the state-of-the-art for centroid-based methods which rely more heavily on pretraining, and use no episodic training~\cite{chen2020new}.

\begin{table}
  \caption{
  \textbf{Effects of architecture and SimCLR Episodes on Prototypical Nets, for Meta-Dataset Train-on-ILSVRC.}  
We ablate architectural choices: use of Exponential Moving Averages (EMA) at test time for Batch Norm (versus computing Batch Norm statistics on the support set at test time), image resolution (224, versus the baseline's 126), ResNet-34 (R34) replacing ResNet-18, SimCLR-style augmentation (SC-Aug), and the addition of 50\% SimCLR Episodes (SC-Eps).
  The test datasets from Meta-Dataset are ImNet: Meta-Dataset's ImageNet Test classes; Omni: OmniGlot drawn characters; Acraft: Aircraft; Bird: CU-Birds; DTD: Textures; QDraw: Quick Draw drawings; Fungi: FGVCx fungi challenge; Flower: VGG Flowers; COCO: Microsoft COCO cropped objects.  
  The best number in each column is bolded, along with others that are within a confidence interval~\cite{metadataset}.  
  $\overline{\text{Rank}}$ is the average rank for each method.  
Using SimCLR Episodes provides improvements on almost all datasets, and provides especially large boosts for datasets which are dissimilar from ImageNet, such as OmniGlot. 
However, simply using SimCLR transformations without instance discrimination (SC-Aug) harms results on almost all datasets.
Increased capacity provides small benefits on some datasets, especially the more realistic and ImageNet-like datasets (e.g.,  birds), but actually harm others like OmniGlot.
Note that in this table, QuickDraw uses the split from the original paper~\cite{metadataset} rather than the (somewhat easier) split published for that paper's public benchmark.  
For all other tables, we use the split from the published benchmark.
}
  \label{tab:simclr-episodes}
  \centering
\addtolength{\tabcolsep}{-3pt}

\resizebox{\textwidth}{!}{
\begin{tabular}{cccccrrrrrrrrrrr}
\toprule
224 & R34 & SC-Aug & SC-Eps & EMA &      ImNet &    Omni &    Acraft &       Bird &    DTD &  QDraw &       Fungi & Flower &     Sign &      COCO &  $\overline{\text{Rank}}$ \\
\midrule
 & & & & \cmark                      &      49.10 &      59.27 &      49.31 &      68.43 &      66.70 &      45.83 &      38.48 &       85.34 &      49.49 &      42.88 &       5.55 \\
 & & & &                            &      49.77 &      55.70 &      52.06 &      68.58 &      67.27 &      49.86 &      37.68 &       84.32 &      50.27 &      41.92 &       5.20 \\
 \cmark & \cmark & & & \cmark        &      51.66 &      57.22 &      51.63 &      71.73 & \bf{69.72} &      47.31 & \bf{42.07} &       87.29 &      47.45 & \bf{44.38} &       4.35 \\
 \cmark & \cmark & & &              & \bf{52.51} &      49.87 &      56.47 & \bf{72.81} &      68.45 &      51.41 & \bf{42.16} &  \bf{87.92} &      54.40 &      40.60 &       3.30 \\
 \cmark & \cmark & \cmark & &  \cmark &     47.58 &      55.73 &      46.93 &      57.75 &      54.88 &      42.91 &      37.42 &       83.82 &      46.88 & \bf{43.36} &       7.55 \\
 \cmark & \cmark & \cmark & &       &      47.94 &      51.79 &      54.58 &      62.84 &      58.64 &      46.36 &      36.06 &       76.88 &      48.35 &      38.77 &       7.45 \\
 \cmark & \cmark & & \cmark & \cmark &      49.67 &      65.21 &      54.46 &      60.94 &      63.96 &      50.64 &      37.84 &  \bf{88.70} &      51.61 & \bf{42.97} &       4.35 \\
 \cmark & \cmark & & \cmark &       & \bf{53.69} & \bf{68.50} & \bf{58.04} & \bf{74.07} & \bf{68.76} & \bf{53.30} & \bf{40.73} &       86.96 & \bf{58.11} &      41.70 &  \bf{1.90} \\
\midrule
\multicolumn{5}{l}{ProtoNets~\cite{metadataset}}         &      50.50 &      59.98 &      53.10 &      68.79 &      66.56 &      48.96 &      39.71 &       85.27 &      47.12 &      41.00 &       5.35 \\
\bottomrule
\end{tabular}
}
\vspace*{-1mm}
\end{table}

\begin{table}
  \caption{
  \textbf{CrossTransformers (CTX) comparison to state-of-the-art.}
  We compare four versions of CrossTransformers to several state-of-the-art methods, which are the best performers among those evaluated for Meta-Dataset Train-on-ILSVRC.  
  We see that CTX alone has better average rank than any baseline.
  Adding SimCLR episodes (+SimCLR Eps) and data augmentation inspired by BOHB~\cite{saikia2020optimized} (+Aug) further improves results.
  Our full model is on-par or above prior methods on all but one dataset, sometimes with large gaps over the best baseline (e.g., +5\% on OmniGlot, +13\% on Aircraft, +5\% on Signs), and furthermore, each prior method has some datasets where we outperform by a larger margin (the next best average rank~\cite{tian2020rethinking}, performs 19\% worse on Aircraft and 17\% worse on OmniGlot).
  Finally, adding a test-time Logistic Regression classifier inspired by Tian et al.~\cite{tian2020rethinking} improves performance on the one dataset---DTD textures---that was otherwise lacking.
  Note that most of these methods~\cite{saikia2020optimized,chen2020new,tian2020rethinking} are unpublished concurrent work.
  }
  \label{tab:crosstransformer}
  \centering
\begin{small}\addtolength{\tabcolsep}{-3pt}
\resizebox{\textwidth}{!}{
\begin{tabular}{lrrrrrrrrrrr}
\toprule
&      ImNet &    Omni &    Acraft &       Bird &    DTD &  QDraw &       Fungi & Flower &     Sign &      COCO &  $\overline{\text{Rank}}$ \\
\midrule
Finetuning~\cite{metadataset}                    &      45.78 &      60.85 &      68.69 &      57.31 &      69.05 &      42.60 &      38.20 &       85.51 &      66.79 &      34.86 &      12.20 \\
ProtoNets~\cite{metadataset}                     &      50.50 &      59.98 &      53.10 &      68.79 &      66.56 &      48.96 &      39.71 &       85.27 &      47.12 &      41.00 &      12.65 \\
ProtoNets+MAML~\cite{metadataset}                &      49.53 &      63.37 &      55.95 &      68.66 &      66.49 &      51.52 &      39.96 &       87.15 &      48.83 &      43.74 &      11.55 \\
CNAPS~\cite{CNAPS}                               &      50.60 &      45.20 &      36.00 &      60.70 &      67.50 &      42.30 &      30.10 &       70.70 &      53.30 &      45.20 &      13.55 \\
BOHB-L~\cite{saikia2020optimized}                &      50.60 &      64.09 &      57.36 &      67.68 &      70.38 &      46.26 &      33.82 &       85.51 &      55.17 &      41.58 &      11.50 \\
BOHB-NC~\cite{saikia2020optimized}               &      51.92 &      67.57 &      54.12 &      70.69 &      68.34 &      50.33 &      41.38 &       87.34 &      51.80 &      48.03 &      10.15 \\
BOHB-NC Ensemble~\cite{saikia2020optimized}      &      55.39 &      77.45 &      60.85 &      73.56 &      72.86 &      61.16 &      44.54 &       90.62 &      57.53 &      51.86 &       7.45 \\
Dhillon et al.~\cite{dhillon2020baseline2}       &          - &          - &      68.69 &      74.26 &      77.35 &          - &          - &       88.14 &      55.98 &      40.62 &       -\\    
Meta-Baseline~\cite{chen2020new}                 &      59.20 &      69.10 &      54.10 &      77.30 &      76.00 &      57.30 &      45.40 &       89.60 &      66.20 &      55.70 &       7.20 \\
Tian et al. LR~\cite{tian2020rethinking}         &      60.14 &      64.92 &      63.12 &      77.69 &      78.59 &      62.48 &      47.12 &       91.60 &      77.51 &      57.00 &       5.50 \\
Tian et al. LR-distill~\cite{tian2020rethinking} &      61.58 &      64.31 &      62.32 &      79.47 & \bf{79.28} &      60.83 &      48.53 &       91.00 &      76.33 & \bf{59.28} &       4.60 \\
ProtoNets (Our implementation)                   &      51.66 &      57.22 &      51.63 &      71.73 &      69.72 &      53.81 &      42.07 &       87.29 &      47.45 &      44.38 &      11.10 \\
\hline
CTX                                              &      61.94 &      76.52 &      79.65 & \bf{84.06} &      76.26 &      65.67 & \bf{52.53} &       94.11 &      70.47 &      53.51 &       3.85 \\
CTX+SimCLR Eps                                   & \bf{63.79} & \bf{80.83} & \bf{82.05} &      82.01 &      75.76 &      68.84 & \bf{52.01} &       94.62 &      75.01 &      52.76 &       3.05 \\
CTX+SimCLR Eps+Aug                               & \bf{62.76} & \bf{82.21} &      79.49 &      80.63 &      75.57 & \bf{72.68} & \bf{51.58} &  \bf{95.34} & \bf{82.65} & \bf{59.90} &  \bf{2.25} \\
CTX+SimCLR Eps+Aug+LR                            &      62.25 & \bf{82.03} &      77.41 &      76.66 & \bf{80.29} & \bf{72.24} &      49.39 &       93.05 &      75.25 & \bf{60.35} &       3.40 \\

\bottomrule
\end{tabular}
}
\end{small}
\vspace*{-4mm}
\end{table}

\subsection{CrossTransformers results}

Given these performant features, we next turn to CrossTransformers.  
\Cref{tab:crosstransformer} compares CrossTransformers (CTX) with and without SimCLR episodes to several state-of-the-art methods, including the Prototypical Nets on which they are based.
We see that CrossTransformers provide strong performance on their own, including having a better average rank than all baselines.
With SimCLR episodes providing more versatile features, we see further improvements, with performance on-par or better than the best methods on almost every dataset.
We note particularly large improvements on OmniGlot, which has a large domain gap relative to the training data.
We also see strong improvements on Street Signs, Aircraft, and Flowers, where multiple test-time categories map to few training-time categories, and often exhibit well-defined spatial correspondence. 

DTD, however, is more challenging for basic CTX, which is unsurprising since textures have little of the kind of spatial correspondence that CTX attempts to find.
COCO is also challenging, likely due to its extremely large intra-class variation (e.g., occlusion) and the fact that many categories overlap with ImageNet-train categories, meaning that simply memorizing categories from the training set may be more useful than using test-time appearance.
To explore this trade-off, we applied logistic regression at test time to a globally pooled feature (see \refsupp{a:impl:global}), which provides additional logits that are averaged with the CTX logits.
We see non-trivial improvements on DTD by using this, although we sacrifice some performance on other datasets, such as Signs and Aircraft.
This implies that there's a fundamental tension between learning categories based on global features, and decomposing the task into local features.  
More work is needed to better combine these two ideas.

Finally, \Cref{fig:attention_viz} depicts the correspondence inferred by the CrossTransformer. The attention is often semantically meaningful: object parts are well matched, including heads, bodies, feet, engines, and strings.
The attention is often not one-to-one either: for the flower, the single query flower is matched to multiple flowers in some of the support images.
Furthermore, the matching works even when the fine-grained classes are not the same, such as the different species of birds, suggesting that the attention is indeed a coarse-grained matching that has not overfit to the training-set classes.

\section{Conclusion}

Within a single domain, deep networks have a remarkable ability to compose and reuse features in order to achieve statistical efficiency.
However, this work shows the hidden problem with such systems: the networks compose features in a way that conflates images which have different appearance but the same label, i.e., it loses information about intra-class variation that may be necessary to understand novel classes. 
We propose two techniques that help resolve this problem: self-supervised learning, which prevents features from losing that intra-class variation, and CrossTransformers, which help neural networks classify images using local features that are more likely to generalize.
However, this problem is far from resolved.
In particular, our algorithm provides less benefit when less spatial structure is available, when knowledge of train-time categories can be useful (as in, e.g., COCO), or when higher-level reasoning is required (e.g., finding conjunctions of multiple objects).
Allowing this algorithm to use spatial structure only where relevant remains an open problem.

\section*{Broader Impact}

The algorithm presented in this paper most directly applies to few-shot recognition, which has numerous uses in industry, including vision systems for robotics that must adapt to new objects, and photo-organizing software which must infer the presence of new classes of objects on-the-fly.
Unfamiliar objects are ubiquitous in many real-world vision applications due to the so-called `long tail'~\cite{wang2017learning} of objects that occur in real scenes, and therefore we expect our algorithm to improve the robustness of visual recognition systems.
While our current work only addresses classification, many other tasks in computer vision, such as object detection and segmentation, use neural network representations that can likewise be made more robust using the kind of architectures presented here.

Our algorithm attempts to build representations which factorize the object recognition problem into sub-problems (feature correspondence and feature comparison) that will each transfer correctly to new datasets.
We hope that further research in this direction may help address dataset biases, including biases regarding race, gender, or other attributes~\cite{yang2020towards}, by helping to disentangle the truly meaningful traits from the spurious correlations.
Finally, while this algorithm presents an advance to state-of-the-art in understanding rare objects, the general performance of such systems is still far below human performance. 
For safety-critical applications (e.g., surgery or self-driving cars), relying on the ability of vision systems to correctly interpret unusual situations is risky with current systems, even with the advances presented here.

\section*{Funding Disclosure}
This work was funded by DeepMind.

\begin{ack}
  The authors would like to thank Pascal Lamblin for help with Meta-Dataset, Olivier Hénaff for help with SimCLR, Yonglong Tian for help in reproducing baselines, and Relja Arandjelović for invaluable advice on the paper.
  They are also grateful to Jean-Baptiste Alayrac, Joao Carreira, Mateusz Malinowski, Viorica Pătrăucean, Adria Recasens, and Lucas Smaira for helpful discussions, support, and feedback on the project.
\end{ack}

{\small
\bibliography{refs/shortstrings,refs/biblio}

\begin{thebibliography}{100}\itemsep=-1pt

\bibitem{andrychowicz16learning}
M.~Andrychowicz, M.~Denil, S.~Gomez, M.~W. Hoffman, D.~Pfau, T.~Schaul,
  B.~Shillingford, and N.~De~Freitas.
\newblock Learning to learn by gradient descent by gradient descent.
\newblock In {\em NeurIPS}, 2016.

\bibitem{bachman19amdim}
P.~Bachman, R.~D. Hjelm, and W.~Buchwalter.
\newblock Learning representations by maximizing mutual information across
  views.
\newblock In {\em NeurIPS}, 2019.

\bibitem{bahdanau2014neural}
D.~Bahdanau, K.~Cho, and Y.~Bengio.
\newblock Neural machine translation by jointly learning to align and
  translate.
\newblock In {\em Proc. ICLR}, 2015.

\bibitem{Bart05}
E.~Bart and S.~Ullman.
\newblock Cross-generalization: Learning novel classes from a single example by
  feature replacement.
\newblock In {\em Proc. CVPR}, 2005.

\bibitem{bengio1992optimization}
S.~Bengio, Y.~Bengio, J.~Cloutier, and J.~Gecsei.
\newblock On the optimization of a synaptic learning rule.
\newblock In {\em Preprints Conf. Optimality in Artificial and Biological
  Neural Networks}, volume~2. Univ. of Texas, 1992.

\bibitem{bengio1990learning}
Y.~Bengio, S.~Bengio, and J.~Cloutier.
\newblock {\em Learning a synaptic learning rule}.
\newblock Citeseer, 1990.

\bibitem{bertinetto2018meta}
L.~Bertinetto, J.~F. Henriques, P.~H. Torr, and A.~Vedaldi.
\newblock Meta-learning with differentiable closed-form solvers.
\newblock In {\em Proc. ICLR}, 2019.

\bibitem{bertinetto2016learning}
L.~Bertinetto, J.~F. Henriques, J.~Valmadre, P.~Torr, and A.~Vedaldi.
\newblock Learning feed-forward one-shot learners.
\newblock In {\em NeurIPS}, 2016.

\bibitem{bourdev2009poselets}
L.~Bourdev and J.~Malik.
\newblock Poselets: Body part detectors trained using 3d human pose
  annotations.
\newblock In {\em Proc. ICCV}, 2009.

\bibitem{bromley1994signature}
J.~Bromley, I.~Guyon, Y.~LeCun, E.~S{\"a}ckinger, and R.~Shah.
\newblock Signature verification using a" siamese" time delay neural network.
\newblock In {\em NeurIPS}, 1994.

\bibitem{bronstein2009partial}
A.~M. Bronstein, M.~M. Bronstein, A.~M. Bruckstein, and R.~Kimmel.
\newblock Partial similarity of objects, or how to compare a centaur to a
  horse.
\newblock {\em Proc. ICCV}, 2009.

\bibitem{cao2020unifying}
B.~Cao, A.~Araujo, and J.~Sim.
\newblock Unifying deep local and global features for image search.
\newblock In {\em Proc. ECCV}, 2020.

\bibitem{caron2018deep}
M.~Caron, P.~Bojanowski, A.~Joulin, and M.~Douze.
\newblock Deep clustering for unsupervised learning of visual features.
\newblock In {\em Proc. ECCV}, 2018.

\bibitem{Chatfield14a}
K.~Chatfield, K.~Simonyan, and A.~Zisserman.
\newblock Efficient on-the-fly category retrieval using convnets and gpus.
\newblock In {\em Asian Conference on Computer Vision}, 2014.

\bibitem{simclr}
T.~Chen, S.~Kornblith, M.~Norouzi, and G.~Hinton.
\newblock A simple framework for contrastive learning of visual
  representations.
\newblock In {\em Proc. ICML}, 2020.

\bibitem{chen2020new}
Y.~Chen, X.~Wang, Z.~Liu, H.~Xu, and T.~Darrell.
\newblock A new meta-baseline for few-shot learning.
\newblock {\em arXiv preprint arXiv:2003.04390}, 2020.

\bibitem{chopra2005learning}
S.~Chopra, R.~Hadsell, and Y.~LeCun.
\newblock Learning a similarity metric discriminatively, with application to
  face verification.
\newblock In {\em Proc. CVPR}, 2005.

\bibitem{cortes2006finite}
J.~Cort{\'e}s.
\newblock Finite-time convergent gradient flows with applications to network
  consensus.
\newblock {\em Automatica}, 42(11), 2006.

\bibitem{autoaugment}
E.~D. Cubuk, B.~Zoph, D.~Mane, V.~Vasudevan, and Q.~V. Le.
\newblock Autoaugment: Learning augmentation strategies from data.
\newblock In {\em Proc. CVPR}, 2019.

\bibitem{dhillon2020baseline2}
G.~S. Dhillon{ }et{ }al.
\newblock A baseline for few-shot image classification.
\newblock {\em Proc. ICLR}, 2020.

\bibitem{doersch15}
C.~Doersch, A.~Gupta, and A.~A. Efros.
\newblock Unsupervised visual representation learning by context prediction.
\newblock In {\em Proc. ICCV}, 2015.

\bibitem{doersch2012what}
C.~Doersch, S.~Singh, A.~Gupta, J.~Sivic, and A.~A. Efros.
\newblock What makes paris look like paris?
\newblock {\em Proc. ACM SIGGRAPH}, 31(4), 2012.

\bibitem{doersch17multi}
C.~Doersch and A.~Zisserman.
\newblock Multi-task self-supervised visual learning.
\newblock In {\em Proceedings of the IEEE International Conference on Computer
  Vision}, pages 2051--2060, 2017.

\bibitem{dosovitskiy14disc}
A.~Dosovitskiy, J.~T. Springenberg, M.~Riedmiller, and T.~Brox.
\newblock Discriminative unsupervised feature learning with convolutional
  neural networks.
\newblock In {\em NeurIPS}. 2014.

\bibitem{fei06one}
L.~Fei-Fei, R.~Fergus, and P.~Perona.
\newblock One-shot learning of object categories.
\newblock {\em IEEE PAMI}, 2006.

\bibitem{felzenszwalb2009object}
P.~F. Felzenszwalb, R.~B. Girshick, D.~McAllester, and D.~Ramanan.
\newblock Object detection with discriminatively trained part-based models.
\newblock {\em IEEE PAMI}, 32(9), 2009.

\bibitem{maml}
C.~Finn, P.~Abbeel, and S.~Levine.
\newblock Model-agnostic meta-learning for fast adaptation of deep networks.
\newblock In {\em Proc. ICML}, 2017.

\bibitem{garcia2017few}
V.~Garcia and J.~Bruna.
\newblock Few-shot learning with graph neural networks.
\newblock In {\em Proc. ICLR}, 2018.

\bibitem{gidaris19fs}
S.~Gidaris, A.~Bursuc, N.~Komodakis, P.~Perez, and M.~Cord.
\newblock Boosting few-shot visual learning with self-supervision.
\newblock In {\em The IEEE International Conference on Computer Vision (ICCV)},
  October 2019.

\bibitem{gidaris18rotation}
S.~Gidaris, P.~Singh, and N.~Komodakis.
\newblock Unsupervised representation learning by predicting image rotations.
\newblock In {\em Proc. ICLR}, 2018.

\bibitem{girshick14rcnn}
R.~Girshick, J.~Donahue, T.~Darrell, and J.~Malik.
\newblock Rich feature hierarchies for accurate object detection and semantic
  segmentation.
\newblock In {\em Proc. CVPR}, 2014.

\bibitem{giusti2013fast}
A.~Giusti, D.~C. Cire{\c{s}}an, J.~Masci, L.~M. Gambardella, and
  J.~Schmidhuber.
\newblock Fast image scanning with deep max-pooling convolutional neural
  networks.
\newblock In {\em Intl. Conf. Image Proc.}, 2013.

\bibitem{ha2016hypernetworks}
D.~Ha, A.~Dai, and Q.~V. Le.
\newblock Hypernetworks.
\newblock {\em arXiv preprint arXiv:1609.09106}, 2016.

\bibitem{han2015matchnet}
X.~Han, T.~Leung, Y.~Jia, R.~Sukthankar, and A.~C. Berg.
\newblock Matchnet: Unifying feature and metric learning for patch-based
  matching.
\newblock In {\em Proc. CVPR}, 2015.

\bibitem{hariharan17low}
B.~Hariharan and R.~Girshick.
\newblock Low-shot visual recognition by shrinking and hallucinating features.
\newblock In {\em Proc. CVPR}, 2017.

\bibitem{moco}
K.~He, H.~Fan, Y.~Wu, S.~Xie, and R.~Girshick.
\newblock Momentum contrast for unsupervised visual representation learning.
\newblock In {\em Proc. CVPR}, 2020.

\bibitem{he2016deep}
K.~He, X.~Zhang, S.~Ren, and J.~Sun.
\newblock Deep residual learning for image recognition.
\newblock In {\em Proc. CVPR}, 2016.

\bibitem{hochreiter01l2l}
S.~Hochreiter, A.~S. Younger, and P.~R. Conwell.
\newblock Learning to learn using gradient descent.
\newblock In {\em International Conference on Artificial Neural Networks}.
  Springer, 2001.

\bibitem{holschneider1990real}
M.~Holschneider, R.~Kronland-Martinet, J.~Morlet, and P.~Tchamitchian.
\newblock A real-time algorithm for signal analysis with the help of the
  wavelet transform.
\newblock In {\em Wavelets}, pages 286--297. 1990.

\bibitem{jacobs2000class}
D.~Jacobs, D.~Weinshall, and Y.~Gdalyahu.
\newblock Class representation and image retrieval with non-metric distances.
\newblock {\em IEEE PAMI}, 22(6):583--600, 2000.

\bibitem{jason2016back}
J.~Y. Jason, A.~W. Harley, and K.~G. Derpanis.
\newblock Back to basics: Unsupervised learning of optical flow via brightness
  constancy and motion smoothness.
\newblock In {\em Proc. CVPR}, 2016.

\bibitem{juneja2013blocks}
M.~Juneja, A.~Vedaldi, C.~Jawahar, and A.~Zisserman.
\newblock Blocks that shout: Distinctive parts for scene classification.
\newblock In {\em Proc. CVPR}, 2013.

\bibitem{kaiser2017learning}
{\L}.~Kaiser, O.~Nachum, A.~Roy, and S.~Bengio.
\newblock Learning to remember rare events.
\newblock In {\em Proc. ICLR}, 2017.

\bibitem{koch2015siamese}
G.~Koch, R.~Zemel, and R.~Salakhutdinov.
\newblock Siamese neural networks for one-shot image recognition.
\newblock In {\em ICML deep learning workshop}, volume~2. Lille, 2015.

\bibitem{kolesnikov2019revisiting}
A.~Kolesnikov, X.~Zhai, and L.~Beyer.
\newblock Revisiting self-supervised visual representation learning.
\newblock In {\em Proc. CVPR}, 2019.

\bibitem{lake15concept}
B.~M. Lake, R.~Salakhutdinov, and J.~B. Tenenbaum.
\newblock Human-level concept learning through probabilistic program induction.
\newblock {\em Science}, 350, 2015.

\bibitem{larsson2016learning}
G.~Larsson, M.~Maire, and G.~Shakhnarovich.
\newblock Learning representations for automatic colorization.
\newblock In {\em Proc. ECCV}, 2016.

\bibitem{li2013probabilistic}
H.~Li, G.~Hua, Z.~Lin, J.~Brandt, and J.~Yang.
\newblock Probabilistic elastic matching for pose variant face verification.
\newblock In {\em Proc. CVPR}, 2013.

\bibitem{lifchitz2019dense}
Y.~Lifchitz, Y.~Avrithis, S.~Picard, and A.~Bursuc.
\newblock Dense classification and implanting for few-shot learning.
\newblock In {\em Proc. CVPR}, 2019.

\bibitem{liu2019selflow}
P.~Liu, M.~Lyu, I.~King, and J.~Xu.
\newblock Selflow: Self-supervised learning of optical flow.
\newblock In {\em Proc. CVPR}, 2019.

\bibitem{maclaurin2015gradient}
D.~Maclaurin, D.~Duvenaud, and R.~Adams.
\newblock Gradient-based hyperparameter optimization through reversible
  learning.
\newblock In {\em International Conference on Machine Learning}, 2015.

\bibitem{maji2013airplanes}
S.~Maji, E.~Rahtu, J.~Kannala, M.~Blaschko, and A.~Vedaldi.
\newblock Fine-grained visual classification of aircraft.
\newblock {\em arXiv preprint arXiv:1306.5151}, 2013.

\bibitem{miller00oneshot}
E.~G. Miller, N.~E. Matsakis, and P.~A. Viola.
\newblock Learning from one example through shared densities on transforms.
\newblock In {\em Proc. CVPR}, 2000.

\bibitem{mishra17snail}
N.~Mishra, M.~Rohaninejad, X.~Chen, and P.~Abbeel.
\newblock A simple neural attentive meta-learner.
\newblock In {\em Proc. ICLR}, 2017.

\bibitem{misra2020self}
I.~Misra and L.~v.~d. Maaten.
\newblock Self-supervised learning of pretext-invariant representations.
\newblock In {\em Proc. CVPR}, 2020.

\bibitem{munkhdalai2017meta}
T.~Munkhdalai and H.~Yu.
\newblock Meta networks.
\newblock In {\em Proc. ICML}, 2017.

\bibitem{naik1992meta}
D.~K. Naik and R.~J. Mammone.
\newblock Meta-neural networks that learn by learning.
\newblock In {\em [Proceedings 1992] IJCNN International Joint Conference on
  Neural Networks}, volume~1. IEEE, 1992.

\bibitem{nesterov1984minimization}
Y.~E. Nesterov.
\newblock Minimization methods for nonsmooth convex and quasiconvex functions.
\newblock {\em Matekon}, 29, 1984.

\bibitem{nichol18first}
J.~Nichol, Alex any Andrychowicz ed~Achiam and J.~Schulman.
\newblock On first-order meta-learning algorithms.
\newblock {\em arXiv preprint arXiv:1803.02999}, 2018.

\bibitem{noroozi2016unsupervised}
M.~Noroozi and P.~Favaro.
\newblock Unsupervised learning of visual representations by solving jigsaw
  puzzles.
\newblock In {\em Proc. ECCV}, 2016.

\bibitem{oquab2014learning}
M.~Oquab, L.~Bottou, I.~Laptev, and J.~Sivic.
\newblock Learning and transferring mid-level image representations using
  convolutional neural networks.
\newblock In {\em Proc. CVPR}, 2014.

\bibitem{perez2018film}
E.~Perez, F.~Strub, H.~De~Vries, V.~Dumoulin, and A.~Courville.
\newblock Film: Visual reasoning with a general conditioning layer.
\newblock In {\em Proc. AAAI}, 2018.

\bibitem{ravi17}
S.~Ravi and H.~Larochelle.
\newblock Optimization as a model for few-shot learning.
\newblock In {\em Proc. ICLR}, 2017.

\bibitem{rebuffi2017learning}
S.-A. Rebuffi, H.~Bilen, and A.~Vedaldi.
\newblock Learning multiple visual domains with residual adapters.
\newblock In {\em NeurIPS}, 2017.

\bibitem{CNAPS}
J.~Requeima, J.~Gordon, J.~Bronskill, S.~Nowozin, and R.~E. Turner.
\newblock Fast and flexible multi-task classification using conditional neural
  adaptive processes.
\newblock In {\em NeurIPS}, 2019.

\bibitem{rocco2018neighbourhood}
I.~Rocco, M.~Cimpoi, R.~Arandjelovi{\'c}, A.~Torii, T.~Pajdla, and J.~Sivic.
\newblock Neighbourhood consensus networks.
\newblock In {\em NeurIPS}, pages 1651--1662, 2018.

\bibitem{imagenet}
O.~Russakovsky, J.~Deng, H.~Su, J.~Krause, S.~Satheesh, S.~Ma, Z.~Huang,
  A.~Karpathy, A.~Khosla, M.~Bernstein, et~al.
\newblock Imagenet large scale visual recognition challenge.
\newblock {\em IJCV}, 115(3), 2015.

\bibitem{saikia2020optimized}
T.~Saikia, T.~Brox, and C.~Schmid.
\newblock Optimized generic feature learning for few-shot classification across
  domains.
\newblock {\em arXiv preprint arXiv:2001.07926}, 2020.

\bibitem{santoro2016meta}
A.~Santoro, S.~Bartunov, M.~Botvinick, D.~Wierstra, and T.~Lillicrap.
\newblock Meta-learning with memory-augmented neural networks.
\newblock In {\em Proc. ICML}, 2016.

\bibitem{savarese2006discriminative}
S.~Savarese, J.~Winn, and A.~Criminisi.
\newblock Discriminative object class models of appearance and shape by
  correlatons.
\newblock In {\em Proc. CVPR}, 2006.

\bibitem{schmidhuber1987evolutionary}
J.~Schmidhuber.
\newblock {\em Evolutionary principles in self-referential learning, or on
  learning how to learn: the meta-meta-... hook}.
\newblock PhD thesis, Technische Universit{\"a}t M{\"u}nchen, 1987.

\bibitem{schmidhuber1992learning}
J.~Schmidhuber.
\newblock Learning to control fast-weight memories: An alternative to dynamic
  recurrent networks.
\newblock {\em Neural Computation}, 4(1):131--139, 1992.

\bibitem{schmidhuber1993neural}
J.~Schmidhuber.
\newblock A neural network that embeds its own meta-levels.
\newblock In {\em IEEE International Conference on Neural Networks}, pages
  407--412. IEEE, 1993.

\bibitem{sivic2003video}
J.~Sivic and A.~Zisserman.
\newblock Video {Google}: A text retrieval approach to object matching in
  videos.
\newblock In {\em Proc. ICCV}, 2003.

\bibitem{protonet}
J.~Snell, K.~Swersky, and R.~Zemel.
\newblock Prototypical networks for few-shot learning.
\newblock In {\em NeurIPS}, 2017.

\bibitem{sprechmann2018memory}
P.~Sprechmann, S.~M. Jayakumar, J.~W. Rae, A.~Pritzel, A.~P. Badia, B.~Uria,
  O.~Vinyals, D.~Hassabis, R.~Pascanu, and C.~Blundell.
\newblock Memory-based parameter adaptation.
\newblock In {\em Proc. ICLR}, 2018.

\bibitem{su2019does}
J.-C. Su, S.~Maji, and B.~Hariharan.
\newblock When does self-supervision improve few-shot learning?
\newblock In {\em Proc. ECCV}, 2020.

\bibitem{relationnet}
F.~Sung, Y.~Yang, L.~Zhang, T.~Xiang, P.~H. Torr, and T.~M. Hospedales.
\newblock Learning to compare: Relation network for few-shot learning.
\newblock In {\em Proc. CVPR}, 2018.

\bibitem{thrun98lifelong}
S.~Thrun.
\newblock Lifelong learning algorithms.
\newblock In {\em Learning to learn}. Springer, 1998.

\bibitem{thrun98learning}
S.~Thrun and L.~Pratt.
\newblock {\em Learning to learn}.
\newblock Springer Science \& Business Media, 1998.

\bibitem{tian2019contrastive}
Y.~Tian, D.~Krishnan, and P.~Isola.
\newblock Contrastive multiview coding.
\newblock {\em arXiv preprint arXiv:1906.05849}, 2019.

\bibitem{tian2020rethinking}
Y.~Tian, Y.~Wang, D.~Krishnan, J.~B. Tenenbaum, and P.~Isola.
\newblock Rethinking few-shot image classification: a good embedding is all you
  need?
\newblock In {\em Proc. ECCV}, 2020.

\bibitem{tokmakov2019learning}
P.~Tokmakov, Y.-X. Wang, and M.~Hebert.
\newblock Learning compositional representations for few-shot recognition.
\newblock In {\em Proc. ICCV}, 2019.

\bibitem{tolias2020learning}
G.~Tolias, T.~Jenicek, and O.~Chum.
\newblock Learning and aggregating deep local descriptors for instance-level
  recognition.
\newblock In {\em Proc. ECCV}, 2020.

\bibitem{metadataset}
E.~Triantafillou, T.~Zhu, V.~Dumoulin, P.~Lamblin, U.~Evci, K.~Xu, R.~Goroshin,
  C.~Gelada, K.~J. Swersky, P.-A. Manzagol, and H.~Larochelle.
\newblock Meta-dataset: A dataset of datasets for learning to learn from few
  examples.
\newblock In {\em Proc. ICLR}, 2020.

\bibitem{oord2016wavenet}
A.~van~den Oord, S.~Dieleman, H.~Zen, K.~Simonyan, O.~Vinyals, A.~Graves,
  N.~Kalchbrenner, A.~Senior, and K.~Kavukcuoglu.
\newblock Wavenet: A generative model for raw audio.
\newblock {\em arXiv preprint arXiv:1609.03499}, 2016.

\bibitem{transformers}
A.~Vaswani, N.~Shazeer, N.~Parmar, J.~Uszkoreit, L.~Jones, A.~N. Gomez,
  {\L}.~Kaiser, and I.~Polosukhin.
\newblock Attention is all you need.
\newblock In {\em NeurIPS}, 2017.

\bibitem{veltkamp2001shape}
R.~C. Veltkamp.
\newblock Shape matching: Similarity measures and algorithms.
\newblock In {\em International Conference on Shape Modeling and Applications},
  2001.

\bibitem{vilalta02perspective}
R.~Vilalta and Y.~Drissi.
\newblock A perspective view and survey of meta-learning.
\newblock {\em Artificial intelligence review}, 18, 2002.

\bibitem{matchnet}
O.~Vinyals, C.~Blundell, T.~Lillicrap, K.~Kavukcuoglu, and D.~Wierstra.
\newblock Matching networks for one shot learning.
\newblock In {\em NeurIPS}, 2016.

\bibitem{vondrick18tracking}
C.~Vondrick, A.~Shrivastava, A.~Fathi, S.~Guadarrama, and K.~Murphy.
\newblock Tracking emerges by colorizing videos.
\newblock In {\em Proc. ECCV}, 2018.

\bibitem{wah2011cubirds}
C.~Wah, S.~Branson, P.~Welinder, P.~Perona, and S.~Belongie.
\newblock {The Caltech-UCSD Birds-200-2011 dataset}.
\newblock 2011.

\bibitem{wang2018non}
X.~Wang, R.~Girshick, A.~Gupta, and K.~He.
\newblock Non-local neural networks.
\newblock In {\em Proc. CVPR}, 2018.

\bibitem{wang2017learning}
Y.-X. Wang, D.~Ramanan, and M.~Hebert.
\newblock Learning to model the tail.
\newblock In {\em NeurIPS}, 2017.

\bibitem{wu2018unsupervised}
Z.~Wu, Y.~Xiong, S.~X. Yu, and D.~Lin.
\newblock Unsupervised feature learning via non-parametric instance
  discrimination.
\newblock In {\em Proc. CVPR}, 2018.

\bibitem{xie18comparator}
W.~Xie, L.~Shen, and A.~Zisserman.
\newblock Comparator networks.
\newblock In {\em Proc. ECCV}, 2018.

\bibitem{yang2020towards}
K.~Yang, K.~Qinami, L.~Fei-Fei, J.~Deng, and O.~Russakovsky.
\newblock Towards fairer datasets: Filtering and balancing the distribution of
  the people subtree in the imagenet hierarchy.
\newblock In {\em Conference on Fairness, Accountability, and Transparency},
  2020.

\bibitem{younger2001meta}
A.~S. Younger, S.~Hochreiter, and P.~R. Conwell.
\newblock Meta-learning with backpropagation.
\newblock In {\em IJCNN'01. International Joint Conference on Neural Networks.
  Proceedings (Cat. No. 01CH37222)}, volume~3. IEEE, 2001.

\bibitem{Zagoruyko2016WRN}
S.~Zagoruyko and N.~Komodakis.
\newblock Wide residual networks.
\newblock In {\em Proc. BMVC.}, 2016.

\bibitem{zhang2019self}
H.~Zhang, I.~Goodfellow, D.~Metaxas, and A.~Odena.
\newblock Self-attention generative adversarial networks.
\newblock In {\em Proc. ICML}, 2019.

\bibitem{zhang2007local}
J.~Zhang, M.~Marsza{\l}ek, S.~Lazebnik, and C.~Schmid.
\newblock Local features and kernels for classification of texture and object
  categories: A comprehensive study.
\newblock {\em IJCV}, 73(2), 2007.

\bibitem{zhang16color}
R.~Zhang, P.~Isola, and A.~A. Efros.
\newblock Colorful image colorization.
\newblock In {\em Proc. ECCV}. Springer, 2016.

\bibitem{zhang2017split}
R.~Zhang, P.~Isola, and A.~A. Efros.
\newblock Split-brain autoencoders: Unsupervised learning by cross-channel
  prediction.
\newblock In {\em Proc. CVPR}, 2017.

\bibitem{zheng2017learning}
H.~Zheng, J.~Fu, T.~Mei, and J.~Luo.
\newblock Learning multi-attention convolutional neural network for
  fine-grained image recognition.
\newblock In {\em Proc. ICCV}, 2017.

\end{thebibliography}
\bibliographystyle{ieee}}

\ifarxiv
\clearpage
\appendix

\section{Out-of-domain vs. within-domain classification implementation details}
In the introduction, we state that algorithms like Prototypical Nets~\cite{protonet} achieve 50\% accuracy on ImageNet held-out categories (out-of-domain), versus 84\% accuracy for a similar challenge given supervised training data (within-domain).

To arrive at these numbers, we used the Meta-Dataset~\cite{metadataset} validation classes.
For Prototypical Nets out-of-domain classification, we find that 50\% is an upper bound: the performance from our reimplementation of Prototypical Nets (using $224{\times}224$ inputs, ResNet 34, and Normalized SGD) on this set is actually 46.4\%. For the supervised within-domain classification baseline we trained a ResNet-34 on the full ImageNet train set (all classes) with $224{\times}224$ images, using modern best practices for training, and found a top-1 performance of 73.0\% on ImageNet's standard val set (i.e., held-out images, but not held-out categories).
We then applied this network to fine-grained classification by constructing query sets, using images from the standard ImageNet val set, but followed the class distribution in Meta-Dataset's query sets for the Meta-Dataset ImageNet val set.
We then classified all these images with the network as following: we discarded all logits except those categories that are present in the query set, in order to ensure that chance performance is the same for the Prototypical Nets and this baseline classifier. The result was 84.2\%.

\section{Supervision Collapse: nearest neighbor experiments}\label{a:nearest_neighbors}
Computing nearest neighbors for Prototypical Net representations proved challenging due to another, entirely different source of supervision collapse: the default implementation of Prototypical Nets actually only produces representations that are comparable \emph{within a single episode}.
This is likely because baseline Prototypical Nets are trained only on episodes: that is, the network only sees fine-grained classification problems (e.g., classifying insects versus other insects), rather than coarse-grained episodes (e.g., insects versus cars).
Therefore, nothing encourages the network to have having distinct, non-overlapping representations of widely different categories (e.g., beetle the insect may have the same representation as beetle the car, without affecting the training loss).
Worse, Batch Norm allows communication within a support set, and so the final representation of each image contains not only information about the image, but also about how it contrasts with other images in the support set.

As a result, even ImageNet images from the Meta-Dataset \emph{training set}, grouped randomly into episodes and fed through prototypical nets, have virtually meaningless representations.
In the file $\operatorname{batch\_norm\_train\_nearest\_neigbors.html}$ in the paper supplement,\footnote{Supplementary material is available on the NeurIPS~2020 webpage for this paper.} we show nearest neighbors retrieved in this way.
The nearest-neighbor retrieval set includes 10\% of images from both the ImageNet train and test sets (Meta-Dataset's split; specifically, $130$ images per class).  
These are passed in batches of size 256 (Batch Norm is set to train mode) to obtain a feature vector for each image.
We use a ResNet-34 Prototypical Net with $224{\times}224$ images, trained with normalized SGD.  
For each row in the HTML file, we show the query image (left), along with the top 9 nearest neighbors, using Euclidean distance.
For $\operatorname{batch\_norm\_train\_nearest\_neigbors.html}$, the results are close to random, even though all query images are taken from the training set.
This is not particularly useful for analysis.

To fix this problem, we make two modifications to Prototypical Net training.  
First, rather than train only on fine-grained episodes, we train on episodes that contain classes sampled uniformly at random from the full ImageNet training set.
This means that a single episode can now contain both cars and insects.
We also replace Batch Norm with Layer Norm, ensuring that there can no longer be communication within the batch.
Results for queries from the training classes are shown in $\operatorname{layer\_norm\_train\_nearest\_neigbors.html}$.
We can see a substantial improvement in the quality of the matches, as would be expected for the retrievals using a representation trained with standard ImageNet classification.

\begin{figure}
  \centering
  \begin{tabular}{ccc}\addtolength{\tabcolsep}{-5pt}
  Same class as query & Any train class & Most frequent train class \\

  \includegraphics[width=.3\textwidth]{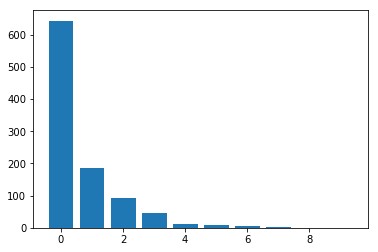} &
  \includegraphics[width=.3\textwidth]{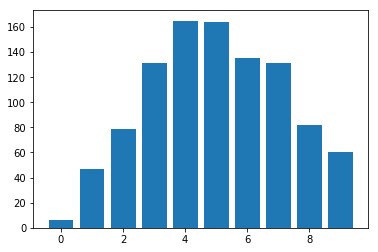} &
  \includegraphics[width=.3\textwidth]{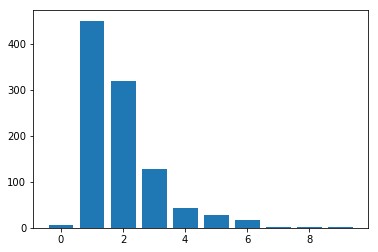} \\
  \end{tabular}
  \caption{\textbf{Nearest neighbors statistics.}
  We sample 1000 random queries from the Meta-Dataset test set, and find the top 9 nearest neighbors in Prototypical Net embedding space, in both training and test sets.
  We show histograms of nearest neighbors: the x-axis is a count of the number of retrievals for a single query that were of some type, and the bar height is the number of queries for which the count was equal to that x-axis value.
  \textbf{Left:} the number of nearest neighbors that came from the same (test) class as the query.
  \textbf{Center:} the number of nearest neighbors that came from the train set.
  \textbf{Right:} the number of retrievals that come from the \emph{same train class}, for the most frequently-retrieved such class.
  Note that the 0'th bin of this plot indicates that all retrievals were from the test set.
  }
  \label{fig:nn_stats}
\end{figure}

Finally, in $\operatorname{layer\_norm\_test\_nearest\_neigbors.html}$, we show the results of using the same retrieval procedure, but using images from test-set classes as queries.
We can see that, due to supervision collapse, the nearest neighbors have returned to being quite poor.

\subsection{Supervision collapse: quantitative analysis}\label{a:collapse:quant}
In \Cref{fig:nn_stats}, we show statistics for a larger set of query images from the test set (1000 queries), which underscore how poor the results are.
In \Cref{fig:nn_stats} left, we see that over 60\% of queries had 0 nearest neighbors of the correct category, even though 130 images of the correct category are guaranteed to exist in the retrieval set.
Furthermore, \Cref{fig:nn_stats} center shows a large proportion of matches for test set images are from the training set.
There are 712 training categories and 130 test categories; therefore, at random chance, we would expect $712/(130+712) = 84.6\%$ to come from the training set (and note that all test-set images are devices, while there are no devices in the train set).
Retrievals from the test set happen more often than chance, but a large fraction do not, and roughly 6\% have \emph{all} retrievals from the training set.
Finally, in \Cref{fig:nn_stats} right, we see that often, the nearest neighbors from the train set are far from random.
Having two or more matches from the the same training set class is quite common (more than 55.3\% of the queries), even more frequent than having even \emph{one} match that's from the correct val set class (34.1\% of the queries).
Many examples have far more than two matches from the same training set class.
In one case, all 9 retrievals were from the same (incorrect) training category.
The statistics reaffirm our intuition that individual Prototypical Net embeddings for held-out images are not likely to capture the correct semantics; instead, the embedding overemphasizes features it has in common with one particular category, which skews its notion of similarity.

We repeated this experiment after training Prototypical Nets with 50\% SimCLR episodes and found improvements: only 43.3\% of queries now have 2 or more neighbors from the same train set category, and 48.8\% have at least one neighbor from the correct class.  
This suggests that SimCLR episodes are effective at reducing supervision collapse, but the problem is far from solved with this technique alone.

\section{CrossTransformer implementation details}\label{a:impl}

\subsection{Training}\label{a:impl:train}

Our experiments with CrossTransformers use no pretraining, although we use it for the experiments involving Prototypical Nets to be consistent with prior work~\cite{metadataset}, which has shown that pretraining gives a boost for Prototypical Net models.  
Specifically, for Prototypical Nets, the representation is pretrained for direct classification on the training set, i.e., the network predicts a fixed number of logits from batches of images sampled uniformly from the training categories.
This is trained with early stopping, where the stopping criterion involves training a linear classifier on validation categories and stopping when this fine-tuning performance begins to decline.
Only then is the network re-architectured into a Prototypical Net, where it is trained on episodes with support sets and query sets with relatively few categories.

For other aspects of training, we follow prior work~\cite{metadataset} where possible, including sampling episodes in the same way, and training the full network using ADAM (applied after normalizing the gradients in the case of Normalized SGD).
We train until convergence, and select the best checkpoint using the error on the validation set.
For the hyperparameters chosen via hyperparameter sweep in the original paper, we use the best values for Prototypical Nets, which are a weight decay of $8.86e-5$, and decaying the learning rate by a factor of $0.915$ after a fixed episode interval.
The exception is the learning rate, where we find that $1.21e-3$ is too high initially, and learning for CTX doesn't take off until it has decayed to half its original value (though this doesn't affect final performance). 
Therefore, we use an initial learning rate of $6e-4$ for all CTX experiments.
For Prototypical Nets, this interval is $500$ episodes, but we use a longer interval for CrossTransformers, as they train from scratch.
For CrossTransformers alone this interval is $2000$; we increase this interval by a factor of two when adding SimCLR Episodes, and another factor of two when adding BOHB-style augmentation, as both of these additions make learning more difficult.

Therefore, the main departures from prior work~\cite{metadataset} are that 1) we use ResNet-34, 2) we feed images at a higher resolution ($224{\times}224$), 3) we use normalized gradient descent, 4) we use 50\% episodes where the categories are selected uniformly at random from ImageNet, 5) we use Batch Norm statistics in test mode at test time (i.e. exponential moving averages computed during training, decaying at a rate of .9 per episode), and 6) we use no pretraining.

\subsection{CrossTransformers architecture}\label{a:impl:ctx}
The output of our ResNet-34 with dilated final block has 512 channels and a $14{\times}14$ grid.
We compute key and value heads with 128 dimensions each, with no non-linearities and no bias.
We find that the attention maps are rather memory-intensive (they contain all pairs of spatial positions between query and support set).
Therefore, we distributed the model across 8 NVIDIA V-100 GPUs, and use gradient rematerialization for the CrossTransformer attention maps.
Training to convergence requires roughly 7 days for our most complex model with SimCLR episodes and BOHB-style augmentations enabled.

\begin{table}
  \caption{
  CrossTransformer comparison of feature map spatial resolution.  
  We see that increasing the spatial resolution from 7 (CTX7) to 14 (CTX14) via dilated convolution typically gives a small performance boost, and almost never harms performance.  
  }
  \label{tab:ctx-res}
  \centering
\begin{small}\addtolength{\tabcolsep}{-3.5pt}
\begin{tabular}{lrrrrrrrrrrr}
\toprule
&      ImNet &    Omni &    Plane &       Bird &    DTD &  QDraw &       Fungi & Flower &     Sign &      COCO &  $\overline{\text{Rank}}$ \\
\midrule
CTX7                   &      59.73 &      74.11 &      70.90 &      80.29 &      73.91 &      65.61 &      48.53 &       91.98 &      68.81 &      50.62 &       5.35 \\
CTX14                                                                             &      61.94 &      76.52 &      79.65 & \bf{84.06} &      76.26 &      65.67 & \bf{52.53} &       94.11 &      70.47 &      53.51 &       3.25 \\
CTX7+SimCLR Eps            &      60.69 &      79.22 &      76.64 &      77.86 & \bf{77.31} &      67.43 &      43.68 &       93.30 &      69.56 &      52.35 &       4.10 \\
CTX14+SimCLR Eps                                                                  & \bf{63.79} &      80.83 & \bf{82.05} &      82.01 &      75.76 &      68.84 & \bf{52.01} &       94.62 &      75.01 &      52.76 &       2.50 \\
CTX7+SimCLR Eps+Aug &      60.76 & \bf{87.26} &      77.56 &      68.31 &      71.44 & \bf{72.62} &      44.12 &       92.45 &      81.20 &      54.66 &       3.85 \\
CTX+SimCLR Eps+Aug                                                              & \bf{62.76} &      82.21 &      79.49 &      80.63 &      75.57 & \bf{72.68} & \bf{51.58} &  \bf{95.34} & \bf{82.65} & \bf{59.90} &  \bf{1.95} \\
\bottomrule
\end{tabular}
\end{small}
\end{table}

To demonstrate the importance of high-resolution feature maps, \Cref{tab:ctx-res} shows the performance of a few versions of CrossTransformer without dilation, which results in stride-32 network with a $7{\times}7$ output grid, like the standard ResNet implementation.
We see that higher resolution almost always gives a small boost.
The boosts are largest on datasets with non-trivial spatial structure where the distinguishing features may be small, e.g., Aircraft, Birds, and Fungi.
On the other hand, the increased resolution makes little difference for DTD textures and QuickDraw, and the the lower resolution actually performs best on OmniGlot.
Textures lack spatial structure, and OmniGlot and Quickdraw contain low-resolution images with few identifying features: therefore, it's unsurprising that the extra resolution isn't useful. 
One possible interpretation is that the network may subdivide the scene into more parts than are justified, which can degrade performance when correspondences are wrong, suggesting that an adaptive mechanism for choosing the resolution might be useful.

\subsection{Augmenting CTX with a global feature}\label{a:impl:global}
Concurrent work~\cite{tian2020rethinking} showed that applying logistic regression to a globally-pooled feature at test time can improve results.
Here, we modify CTX to use the same ideas.
We first globally pool the feature $\Phi(x)$ spatially, which results in a flat 512-dimension vector for each image in both the query and test set.
We then train a simple Logistic Regression classifier using the same parameters from \cite{tian2020rethinking} (sklearn's implementation with a multinomial loss $C{=}10$, applied to $\ell_2$-normalized features).  
Running the classifier on the query images produces another set of logits, which we find are scaled smaller than CTX logits.
Therefore, we produce final classifier logits via $\operatorname{argmax}(\operatorname{CTX}(S,x^q)+\lambda \operatorname{LR}(S,x^q))$, where $\operatorname{CTX}(S,x^q)$ is the logits produced by CTX for query $x^q$ and support set $S$, $\operatorname{LR}(S,x^q)$ are the logits from the logistic regression classifier, and $\lambda$ is a scalar constant that we set to $5$.

We find that this provides no benefit if the embedding network is trained purely as a CrossTransformer.  
However, we find benefits on some datasets (notably DTD) if we add an auxiliary loss that matches the loss used in concurrent work~\cite{tian2020rethinking}.
That is, we compute $\Phi(x)$ for each image $x$ in the support set, and globally pool this feature.  
We then apply a fixed classifier on top of this feature, which performs the 712-way classification for the 712 ImageNet-train categories.
We add this classification loss to the CTX loss (without weight) at training time.

\subsection{Augmentation}\label{a:impl:aug}
For most experiments in the paper, we use no augmentation for images that aren't a part of SimCLR Episodes, following prior work~\cite{metadataset}.
However, BOHB~\cite{saikia2020optimized} studies augmentation extensively, and so we adopt similar augmentation for some experiments.
BOHB optimizes parameters for augmentation in a similar style to AutoAugment~\cite{autoaugment}, using 2 randomly-selected stages, where each stage may consist of rotation, posterizing, solarizing, color shifts, contrast, brightness, sharpness, shear, translation, and cutout, with settings discovered via validation on Meta-Dataset's ImageNet val split.  
We use these only for experiments labeled as ``+Aug'' in~\Cref{tab:crosstransformer} and~\Cref{tab:ctx-res}.
We found qualitatively, however, that even with these changes, the network could still be quite sensitive to input images which are resized from very small images, especially when the input resolution is high ($224{\times}224$ in our case).
Therefore, we add one more stage that the augmentation function can select, which randomly resizes the image by a ratio sampled uniformly from 1 all the way down to a ratio that would produce a 10-pixel-wide image.
Then we compress the image with jpeg, with a compression quality uniformly sampled between 75 and 100, before decompressing and resizing to the original resolution.
These parameters were chosen once, and we ran no hyperparameter sweep to tune them.
We expect that properly tuning them on validation data following BOHB~\cite{saikia2020optimized} would yield further improvements, but we leave this for future work.

Our implementation of SimCLR follows the public one released by the original authors, using the standard two augmentation ops of random cropping and color jittering with the default parameters.
Also following this implementation, we apply random blur to only one image in each pair of positive matches; in our case, we apply the blur to the query image.

\section{Correspondence visualization}\label{a:corrviz}
Figures~\ref{acraft}--\ref{quickdraw} visualize the attention inferred by CrossTransformers for all the 10 evaluation datasets in Meta-Dataset. We show query images for each dataset, along with three support-set images for each. Within each query image, we choose three spatial locations (red, green, and blue squares), and plot the CrossTransformer attention weights for each one in the corresponding color (higher weight means brighter colors). Both $7{\times}7$ and $14{\times}14$ attention maps corresponding to CTX7 and CTX14 models respectively are presented, with the query points selected at approximately the same location for both models.
The inferred correspondences are semantically meaningful, and often not one-to-one.

Finally, \Cref{fig:failure_viz} shows some qualitative examples of a few challenging cases which suggests areas for improvement.  
In particular, our method sometimes produces confident correspondences even when the true correspondence is unclear; in such cases, we might prefer that the method falls back to global comparisons.
Conversely, the algorithm may not always find correspondences when they are available, if, for example, there is a large difference in appearance between corresponding points. 
This suggests that the correspondence itself may be overfitting, and suggests a possible avenue for future research.

\begin{figure}
  \centering
  \addtolength{\tabcolsep}{-4pt}
  \centerline{%
  \resizebox{\linewidth}{!}{%
    \begin{tabular}{cc|c|c|c|c|c}
      \rotatebox{90}{\parbox[t][][t]{2cm}{\centering \small Query}} &
      \includegraphics[height=20mm]{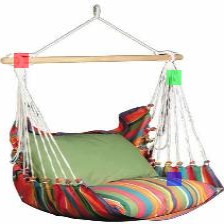} & 
      \includegraphics[height=20mm]{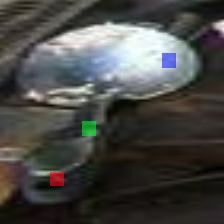} & 
      \includegraphics[height=20mm]{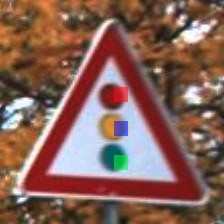} & 
      \includegraphics[height=20mm]{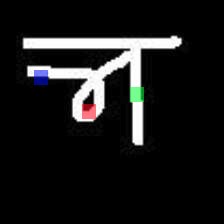} & 
      \includegraphics[height=20mm]{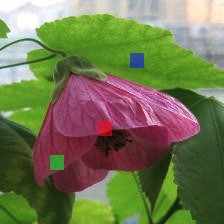} & 
      \includegraphics[height=20mm]{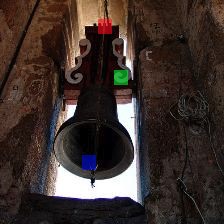} \\
      \rotatebox{90}{\parbox[t][][t]{2cm}{\centering \small Correspondence\\in support set}} &
      \includegraphics[height=20mm]{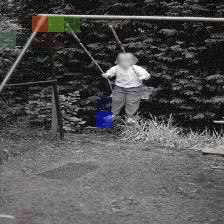} & 
      \includegraphics[height=20mm]{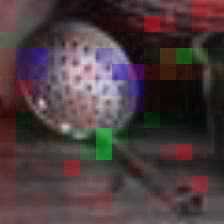} & 
      \includegraphics[height=20mm]{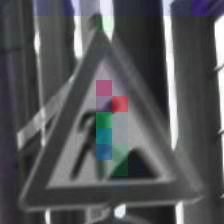} & 
      \includegraphics[height=20mm]{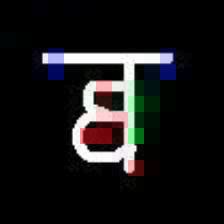} & 
      \includegraphics[height=20mm]{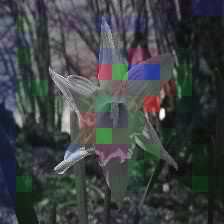} & 
      \includegraphics[height=20mm]{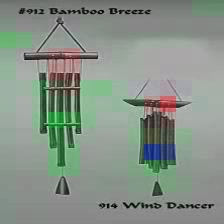} \\
       & {\small (1)} & {\small (2)} & {\small (3)} & {\small (4)} & {\small (5)} & {\small (6)}
    \end{tabular}}}
  \caption{\textbf{Failure cases.} We visualize the attention correspondences for a few challenging cases for our method. In (1) incorrect parts of the swings are matched with high confidence. In the (2) the handle of the ladle is not localized. A shortcoming of our approach is that the model is tasked to match images which cannot be/are difficult to put into correspondence. For example, in (3) and (4) no clear correspondence is feasible, and the model matches parts which are plausibly in the same geometric location.
  Finally, in (5) and (6), the model localizes the instances correctly, but due to large variance in the instance shapes fails at detailed matching of the sub-parts.}
  \label{fig:failure_viz}
\end{figure}

\section{Five-shot}
We also report five-shot results. 
As the standard Meta-Dataset evaluation specifies a broad range of possible shots (e.g., up to 100 examples in extremely rare cases), we believe that 5-shot results can aid in interpretability.
The `ways' are sampled as before (i.e., the standard for Meta-Dataset), but the number of examples per category is set to exactly five.
This means that both support set and query set are class balanced (unlike the standard evaluation, where the query set is balanced but the support set is not).
We use the same checkpoints that were used above (i.e., no additional validation to choose a checkpoint for five-shot evaluation).
Results are shown in \Cref{t:five_shot}.
We see that the performance is somewhat lower, but the overall trends of performance on the datasets are similar.

\section{Confidence intervals}
\Cref{t:ci_table} shows confidence intervals for most experiments in this paper, to enable future comparisons like the tables shown in this work.

\clearpage
\newgeometry{noheadfoot=true,top=0cm,bottom=0mm,left=2cm,right=2cm}
\null\vfill\vfill
\attnviz{acraft}{Aircraft}{Various aircraft parts (e.g., wings, head, tail, engine, landing wheels) are 
matched across instances with large differences in viewpoint/pose and scale}{21mm}
\vfill\vspace{-3mm}
\attnviz{birds}{CU-Birds}{Beak, body, and feet are matched for different species}{21mm}
\vfill\vfill
\restoregeometry

\clearpage
\newgeometry{noheadfoot=true,top=0cm,bottom=0mm,left=2cm,right=2cm}
\null\vfill\vfill
\attnviz{texture}{Describable Textures (DTD)}{Textures do not have localized parts/sub-parts, and hence, the correspondence is quite diffuse (e.g., the first above).  However, specific features, if present, are matched: in the last four examples, the donut's sprinkles and icing, the net-like pattern, the facial features, and the pumpkin's stem, surface and cavities, are matched respectively.}{21mm}%
\vfill
\attnviz{fungi}{FGVCx Fungi}{The caps and stem are matched across variations in pose/rotation, and also number of fungi exhibit one-to-many matches.}{21mm}
\vfill\vfill
\restoregeometry

\clearpage
\newgeometry{noheadfoot=true,top=0cm,bottom=0mm,left=2cm,right=2cm}
\null\vfill\vfill
\attnviz{flowers}{VGG Flower}{The central disk and petals are matched with all instances of flowers present in the support-set images. The background is diffuse and separated from the central object}{21mm}
\vfill\vspace{-3mm}
\attnviz{imnet}{ImageNet}{Correspondence is established in the presence of distractors and large variations in object pose and shape}{21mm}
\vfill\vfill
\restoregeometry

\clearpage
\newgeometry{noheadfoot=true,top=0cm,bottom=0mm,left=2cm,right=2cm}
\null\vfill\vfill
\attnviz{mscoco}{MSCOCO}{We observe some detailed matching for different object categories: animal parts (first two), giraffe ossicones (first), bus, large variations in the pose of rackets (fourth), and different letters in the traffic sign (last)}{21mm}
\vfill\vspace{-3mm}
\attnviz{omni}{Omniglot}{Corresponding parts of various character glyphs are matched}{21mm}
\vfill\vfill
\restoregeometry

\clearpage
\newgeometry{noheadfoot=true,top=0cm,bottom=0mm,left=2cm,right=2cm}
\null\vfill\vfill
\attnviz{signs}{Traffic Signs}{Corresponding parts are matched, even when the support-image is flipped horizontally (last)}{21mm}
\vfill\vspace{-3mm}
\attnviz{quickdraw}{Quick Draw}{Matches across deformations of doodles are observed}{21mm}
\vfill\vfill
\restoregeometry

\clearpage

\newgeometry{top=1in,bottom=1in,left=1cm,right=1cm}
\begin{landscape}
\begin{table}
  \centering
  \caption{
    \textbf{Five-Shot.} For interpretability, we also compute 5-shot results for all Meta-Dataset datasets, and include (accuracy (\%) $\pm$ confidence (\%)) for the sake of future comparison for CTX models (with the 14x14 feature grid).  These are the standard confidence intervals computed for Meta-Dataset: i.e., the standard error computed from the episode-to-episode variability in accuracy across 600 test episodes.}\vspace{2mm}
  \label{t:five_shot}
  \addtolength{\tabcolsep}{-2pt}
  \renewcommand{\arraystretch}{2.5}
  \resizebox*{23cm}{!}{%
  \begin{tabular}{lrrrrrrrrrrr}
    \toprule
& \multicolumn{1}{c}{ILSVRC} & \multicolumn{1}{c}{Omniglot} & \multicolumn{1}{c}{Aircraft} & \multicolumn{1}{c}{Birds} & \multicolumn{1}{c}{Textures} & \multicolumn{1}{c}{Quick Draw} & \multicolumn{1}{c}{Fungi} & \multicolumn{1}{c}{VGG Flowers} & \multicolumn{1}{c}{Traffic} & \multicolumn{1}{c}{MSCOCO}  & \multicolumn{1}{c}{$\overline{\text{Rank}}$} \\ \midrule
Prototypical (ours)   &      41.87$\pm$0.89 &      61.33$\pm$1.13 &      39.40$\pm$0.78 &      65.57$\pm$0.73 &      59.06$\pm$0.60 &      47.86$\pm$0.80 &      41.64$\pm$1.02 &      83.88$\pm$0.48 &      44.84$\pm$0.88 &      41.14$\pm$0.82 &       4.00 \\
CTX                   & \bf{51.70}$\pm$0.90 &      84.24$\pm$0.79 &      62.29$\pm$0.73 & \bf{79.38}$\pm$0.54 & \bf{65.86}$\pm$0.58 &      63.36$\pm$0.73 & \bf{49.43}$\pm$0.98 &      92.74$\pm$0.29 &      68.31$\pm$0.71 &      48.63$\pm$0.79 &       2.25 \\
CTX+SimCLR Eps        &      51.29$\pm$0.89 &      86.14$\pm$0.74 & \bf{69.74}$\pm$0.67 &      74.85$\pm$0.62 &      63.84$\pm$0.62 &      64.11$\pm$0.67 & \bf{48.87}$\pm$0.91 & \bf{93.00}$\pm$0.30 &      70.62$\pm$0.68 &      48.45$\pm$0.83 &       2.10 \\
CTX+SimCLR Eps+Aug & \bf{52.56}$\pm$0.86 & \bf{87.53}$\pm$0.61 &      64.28$\pm$0.71 &      73.27$\pm$0.63 &      64.72$\pm$0.63 & \bf{66.90}$\pm$0.66 & \bf{48.22}$\pm$0.94 & \bf{93.23}$\pm$0.28 & \bf{78.45}$\pm$0.60 & \bf{56.61}$\pm$0.78 &  \bf{1.65} \\
\bottomrule
  \end{tabular}}
\end{table}

\begin{table}
  \centering
  \caption{
    \textbf{Confidence intervals for quantitative results.} We report the confidence intervals in addition to the mean accuracy (accuracy (\%) $\pm$ confidence (\%)) for the models introduced in this work for the sake of future comparison.  All versions shown here use 224 resolution, ResNet34, and exponential moving average (EMA) for test-time batch norm (BN), unless otherwise specified.}\vspace{2mm}
  \label{t:ci_table}
  \addtolength{\tabcolsep}{-2pt}
  \renewcommand{\arraystretch}{2.5}
  \resizebox*{23cm}{!}{%
  \begin{tabular}{lcccccrrrrrrrrrr}
    \toprule
& \multicolumn{1}{c}{ILSVRC} & \multicolumn{1}{c}{Omniglot} & \multicolumn{1}{c}{Aircraft} & \multicolumn{1}{c}{Birds} & \multicolumn{1}{c}{Textures} & \multicolumn{1}{c}{Quick Draw} & \multicolumn{1}{c}{Fungi} & \multicolumn{1}{c}{VGG Flowers} & \multicolumn{1}{c}{Traffic} & \multicolumn{1}{c}{MSCOCO} \\ \midrule

ProtoNets (ours)                                                 &      51.66$\pm$1.10 &      57.22$\pm$1.34 &      51.63$\pm$0.93 &      71.73$\pm$1.02 &      69.72$\pm$0.76 &      53.81$\pm$1.04 &      42.07$\pm$1.14 &      87.29$\pm$0.69 &      47.45$\pm$0.92 &      44.38$\pm$1.03 \\
ProtoNets (ours)+SimCLR Eps                                             &      49.67$\pm$1.06 &      65.21$\pm$1.23 &      54.46$\pm$0.91 &      60.94$\pm$0.94 &      63.96$\pm$0.77 &      50.64$\pm$1.05 &      37.84$\pm$1.06 &      88.70$\pm$0.60 &      51.61$\pm$1.00 &      42.97$\pm$1.04 \\
ProtoNets (ours)+SimCLR Eps (no BN EMA)                                                 &      53.69$\pm$1.10 &      67.44$\pm$1.26 &      57.10$\pm$0.99 &      74.07$\pm$0.91 &      69.46$\pm$0.75 &      51.83$\pm$1.02 &      41.67$\pm$1.21 &      86.93$\pm$0.65 &      57.41$\pm$1.04 &      41.43$\pm$1.10 \\
CTX7                   &      59.73$\pm$1.08 &      74.11$\pm$1.22 &      70.90$\pm$0.99 &      80.29$\pm$0.86 &      73.91$\pm$0.70 &      65.61$\pm$0.82 &      48.53$\pm$1.09 &      91.98$\pm$0.52 &      68.81$\pm$0.99 &      50.62$\pm$1.03 \\
CTX14                                                                          &      61.94$\pm$1.04 &      76.52$\pm$1.14 &      79.65$\pm$0.91 & 84.06$\pm$0.85 &      76.26$\pm$0.70 &      65.67$\pm$0.91 & 52.53$\pm$1.16 &      94.11$\pm$0.44 &      70.47$\pm$0.92 &      53.51$\pm$1.06 \\
CTX7+SimCLR Eps             &      60.69$\pm$0.99 &      79.22$\pm$1.16 &      76.64$\pm$0.88 &      77.86$\pm$0.92 &      77.31$\pm$0.66 &      67.43$\pm$0.88 &      47.37$\pm$1.15 &      93.30$\pm$0.43 &      69.56$\pm$0.99 &      52.35$\pm$1.01 \\
CTX14+SimCLR Eps                                                                   & 63.79$\pm$1.00 &      80.83$\pm$1.07 & 82.05$\pm$0.83 &      82.01$\pm$0.89 &      75.76$\pm$0.76 &      68.84$\pm$0.88 & 52.01$\pm$1.13 &      94.62$\pm$0.43 &      75.01$\pm$0.93 &      52.76$\pm$1.03 \\
CTX7+SimCLR Eps+Aug  &      61.20$\pm$1.04 & 87.26$\pm$0.65 &      77.98$\pm$0.89 &      68.31$\pm$0.71 &      72.70$\pm$0.71 & 73.32$\pm$0.77 &      44.12$\pm$0.94 &      93.29$\pm$0.43 &      80.03$\pm$0.80 &      57.88$\pm$1.04 \\
CTX14+SimCLR Eps+Aug                                                                & 62.76$\pm$0.99 &      82.21$\pm$1.00 &      79.49$\pm$0.89 &      80.63$\pm$0.88 &      75.57$\pm$0.64 & 72.68$\pm$0.82 & 51.58$\pm$1.11 & 95.34$\pm$0.37 & 82.65$\pm$0.76 & 59.90$\pm$1.02 \\
CTX14+SimCLR Eps+Aug+LR                                                          &      62.25$\pm$0.96 &      82.03$\pm$0.98 &      77.41$\pm$0.84 &      76.66$\pm$0.87 & 80.29$\pm$0.72 & 72.24$\pm$0.81 &      49.39$\pm$1.17 &      93.05$\pm$0.50 &      75.25$\pm$0.93 & 60.35$\pm$1.06 \\
\bottomrule
  \end{tabular}}
\end{table}
\end{landscape}
\fi

\end{document}